\documentclass{article}

\usepackage[preprint]{neurips_2026}

\usepackage[utf8]{inputenc}
\usepackage[T1]{fontenc}

\usepackage{amsmath, amsfonts, amssymb}
\usepackage{placeins}
\usepackage{graphicx}
\usepackage{booktabs}
\usepackage{multirow}
\usepackage{makecell}
\usepackage{subcaption}
\usepackage{xcolor}
\usepackage{algorithm}
\usepackage{algorithmic}
\usepackage[breaklinks=true,colorlinks=false,hidelinks]{hyperref}

\newcommand{\prm}{\textsc{PRM}}

\newcommand{\Wt}[1]{W_{t,#1}}
\newcommand{\qt}{q_t}
\newcommand{\dt}{\delta_t}
\newcommand{\at}{\alpha_t}
\newcommand{\Fbar}{\bar{F}}

\title{When the Merge Coefficient Stops Mattering:\\
Proximity Regularized Merging for Continual LoRA Adaptation}
\author{%
  Yixuan Liu\\
  Sapient Intelligence\\
  Tsinghua University\\
  \And
  Yuhao Sun \\
  Sapient Intelligence\\
  \And
  Sen Song$^{\dagger}$ \\
  Tsinghua University\\
  \And
  Jin Li$^{\dagger}$ \\
  Sapient Intelligence\\
  \And
}
\date{}

\begin{document}

\maketitle

\begin{abstract}
Rehearsal-free continual learning with parameter-efficient adapters can be cast as a sequence of task-vector write-in operations: for each new task, a low-rank adapter is learned and merged into a running model.
We propose Proximity Regularized Merging (PRM), a minimal modification to
sequential LoRA merging that adds a proximal penalty during task-vector
training without changing the subsequent write-in rule.
PRM acts as a robust task-vector regularizer: in the reported
Base$\rightarrow$+Prox diagnostics, it improves AAA across
multiple write-in rules, backbones, and class-incremental settings, while its fixed-coefficient variant remains competitive with strong coefficient-based baselines. Mechanistically, matched-prefix norm controls and proximal-strength sweeps show that proximal training shrinks the task-vector radius, lowers Fisher-weighted interference, broadens the coefficient plateau,
and exposes a stability--plasticity trade-off.
Together, these results suggest that the effectiveness of sequential LoRA merging depends not only on how much of a task vector is written in, but also on whether the task vector itself has been trained to be mergeable.
\end{abstract}

\section{Introduction}
\label{sec:intro}

Modern pretrained models are increasingly expected to acquire new
classes, domains, and tasks after deployment. Continual learning studies
this problem under a sequential data regime, where a model must learn
from the current task while retaining performance on previous ones. In
the rehearsal-free setting, the constraint is sharper: previous task data
cannot be stored or replayed because of privacy, storage, or deployment
constraints. Existing rehearsal-free approaches address this challenge
from different angles. Parameter-isolation and prompt-based methods
reduce interference by routing computation through task-specific
components or learned prompts~\citep{wang2022l2p,wang2022dualprompt,smith2023codaprompt},
while regularization-based methods constrain changes to parameters
estimated to be important for previous tasks~\citep{kirkpatrick2017ewc,zenke2017si,aljundi2018mas}.
These approaches have substantially improved rehearsal-free continual
learning, but they leave open a complementary consolidation question:
when a new task is represented as a compact parameter update, under what
conditions can that update be safely absorbed into the single running
model that must serve all tasks seen so far?

Task vectors and model merging provide a natural language for this
question. In model merging, fine-tuned parameter differences are treated
as task vectors and combined to reuse capabilities across
models~\citep{ilharco2023task,yadav2023ties,yu2024dare}. With
LoRA-style parameter-efficient adaptation, each new task produces a
compact low-rank update that can be stored, scaled, and merged into the
current model~\citep{hu2022lora}. In sequential LoRA merging, it is
important to distinguish two objects that are often coupled in practice:
the task vector and the merge coefficient. The task vector, denoted
informally as $\delta_t$, is the update learned from the current task and
therefore determines the direction and scale of the candidate model
change. The merge coefficient, denoted $\alpha_t$, is a scalar used only
when consolidating that learned vector into the running model, e.g., by
forming $\theta_{t-1}+\alpha_t\delta_t$. Recent rehearsal-free
continual-merging methods instantiate this idea with different vector
training procedures and coefficient rules, including MagMax, BECAME, and
P\&M~\citep{marczak2024magmax,li2025became,qiu2025pm}. In particular,
P\&M highlights the coefficient as a central design object: after
perturbation-aware LoRA training, it computes a Fisher-based coefficient
to decide how much of the learned vector should be written into the
running model.

This coefficient-centric view is useful, but incomplete. A merge
coefficient can scale a learned vector, but it cannot change the path
induced by that vector. If useful performance occurs only in a narrow
interval of $\alpha_t$, then consolidation remains fragile: approximation
noise in the coefficient estimate, task-order variation, or a mismatch
between validation and test behavior can move the model away from the
safe part of the path. Conversely, when the learned task vector already
induces a stable write-in path, many coefficients lead to similar
seen-task performance, and the exact coefficient rule becomes less
important. Thus, the central question is not whether a task vector is
``mergeable'' in the abstract, nor only how to tune $\alpha_t$ for a
fixed vector. We instead ask: \textbf{under what training conditions can
LoRA task vectors be merged robustly into the running model, with weaker
forgetting and reduced dependence on the merge coefficient?}

Our method is motivated by a simple observation shared by several lines
of prior work: forgetting is closely related to where and how far the
model moves in parameter space. Classical continual-learning
regularizers penalize changes to parameters estimated to matter for past
tasks~\citep{kirkpatrick2017ewc,zenke2017si,aljundi2018mas};
Fisher-based merging uses curvature information to avoid destructive
combinations~\citep{matena2022merging}; and flatness, weight averaging,
and mode-connectivity studies show that endpoint accuracy alone is not
enough---the path between models also matters~\citep{wortsman2022model,foret2021sam,izmailov2018swa,mirzadeh2021lmc}.
These observations suggest that a LoRA task vector should not merely fit
the current task; it should also remain close enough to the current model
so that scaling it by different coefficients does not quickly enter
regions that interfere with previous tasks.

In this work, we propose \prm{} (\emph{Proximity Regularized Merging}),
which adds a proximal penalty to the current task's LoRA factors during
task-vector training while leaving the downstream write-in rule unchanged.
This lets us test whether robust continual merging is driven less by finding a
delicate coefficient for a fixed vector, and more by learning a vector whose
write-in path is already stable. Across fixed-$\alpha$, CoFiMA, CoMA,
Model-Avg, BECAME, and MagMax~\citep{marczak2024magmax,li2025became,marouf2024weighted},
proximal shaping consistently improves the corresponding base rule. With one
fixed coefficient, \prm{} attains the best reported mean \textsc{AAA} across
our tested settings, indicating that the gain is not merely better coefficient
selection. Mechanism analyses show that PRM mainly shrinks the task-vector
radius, lowering a norm-based bound on Fisher-weighted interference and
broadening the coefficient plateau, until excessive regularization harms
current-task plasticity. These findings support a simple conclusion: training
LoRA task vectors to be easier to merge can weaken forgetting and make precise
merge-coefficient selection less critical.

\section{Related Work}
\label{sec:related}

\paragraph{Rehearsal-free continual learning.}
Rehearsal-free continual learning studies how to learn a sequence of
tasks without storing previous-task data. Traditional methods reduce
forgetting through parameter regularization and importance estimation,
such as EWC, SI, and MAS
\citep{kirkpatrick2017ewc,zenke2017si,aljundi2018mas};
output distillation, as in LwF \citep{li2016lwf}; or gradient and
subspace projection, as in GPM \citep{saha2021gpm}. For pretrained
models, prompt-based approaches learn task-adaptive prompts or prompt
components, including L2P, DualPrompt, and CODA-Prompt
\citep{wang2022l2p,wang2022dualprompt,smith2023codaprompt}. Parameter-efficient methods instead update compact modules or low-rank parameters:
LoRA provides a widely used low-rank adaptation parameterization
\citep{hu2022lora}, and recent LoRA-based continual learners such as
InfLoRA, SD-LoRA, and CL-LoRA further adapt low-rank modules for
rehearsal-free class-incremental learning
\citep{liang2024inflora,wu2024sdlora,he2025cllora}. In this work, we use
LoRA task vectors as the update representation for sequential write-in.

\paragraph{Task vectors and model merging.}
Model merging combines the parameters or task vectors of separately
trained models, for example through weight averaging and model soups
\citep{wortsman2022model}, Fisher-weighted averaging
\citep{matena2022merging}, task arithmetic \citep{ilharco2023task},
or interference-aware merging rules such as TIES and DARE
\citep{yadav2023ties,yu2024dare}. These methods are typically
studied in a static setting where all source models or task vectors are
available at merge time. Continual learning requires a sequential
version of this problem: each new update must be incorporated before the
next task arrives. Continual model-averaging methods such as CoMA and
CoFiMA combine the previous and current task models, with CoFiMA using
Fisher information to weight parameters during averaging
\citep{marouf2024weighted}. MagMax performs sequential task-vector
merging through maximum-magnitude update selection
\citep{marczak2024magmax}, while BECAME derives an adaptive merging
coefficient from a Bayesian continual-learning formulation
\citep{li2025became}. P\&M further formulates a two-stage sequential merging procedure, training a perturbation-aware task-vector proposal and then computing a Fisher-derived write-in coefficient for consolidating it into the running model~\citep{qiu2025pm}. \prm{} follows this sequential LoRA write-in template but modifies only the task-vector training objective, adding a proximal term before any downstream write-in rule is applied.
\section{Method: Shaping LoRA Task Vectors for Mergeability}
\label{sec:method}

We study rehearsal-free class-incremental learning with a frozen
pretrained backbone and sequentially learned LoRA task vectors. Our goal
is to learn task vectors that are not only accurate on the current task,
but also easy to write into the running model. We first formulate
sequential LoRA write-in, then motivate \prm{} as a proximal
task-vector shaping objective, and finally define the mergeability
metrics used in the analysis.

\subsection{Sequential LoRA Write-in}
\label{sec:method:setup}

A task sequence
\(\mathcal{T}_1,\dots,\mathcal{T}_T\) arrives one task at a time.
When training task \(t\), examples from previous tasks
\(\mathcal{T}_{<t}\) are unavailable. Let \(\theta_{t-1}\) denote the
running model before task \(t\). For each adapted projection, fresh LoRA
factors \(A_t,B_t\) induce an effective task vector
\begin{equation}
    \dt = s_{\mathrm{LoRA}} B_t A_t ,
    \label{eq:effective-task-vector}
\end{equation}
where \(s_{\mathrm{LoRA}}\) is the LoRA scaling factor. In practice,
\(\dt\) denotes the collection of such low-rank updates across all
adapted projections. After task-vector training on the current task, this
task vector is written into the running model by
\begin{equation}
    \theta_t = \theta_{t-1} + \at \dt,
    \qquad \at \in [0,1].
    \label{eq:sequential-writein}
\end{equation}

This one-dimensional write-in operation separates two design questions:
which task vector should be learned, and which coefficient should be
used to write it into the running model? Existing coefficient-centric
write-in rules mainly address the second question: given a learned task
vector, how much of it should be written into the running model?
Fisher-derived rules such as P\&M~\citep{qiu2025pm} are one instance of
this view. In contrast, \prm{} focuses on the first question: can
task-vector training itself make the induced write-in path more
mergeable, so that performance becomes less sensitive to the precise
choice of \(\at\)?

\subsection{Mergeability and Write-in Curvature}
\label{sec:method:mergeability}

We call \(\dt\) mergeable if this path is stable across a broad range
of \(\alpha\) coefficients while retaining current-task plasticity. A mergeable task
vector should achieve high peak performance at some
coefficient, but should not require a narrowly tuned coefficient to be
usable.

To motivate the objective, consider the old-task loss
\(\mathcal{L}_{<t}\) along the write-in path. A local second-order
expansion around \(\theta_{t-1}\) gives
\begin{equation}
\mathcal{L}_{<t}(\theta_{t-1}+\alpha\dt)
\approx
\mathcal{L}_{<t}(\theta_{t-1})
+
\alpha g_{<t}^{\top}\dt
+
\frac{1}{2}\alpha^2 \dt^{\top}H_{<t}\dt .
\label{eq:old-loss-expansion-main}
\end{equation}
The linear term can shift the preferred coefficient, while the quadratic
term controls how sharply the old-task loss changes along the write-in
path. Thus, the curvature-sensitive component of coefficient
sensitivity is governed by \(\dt^{\top}H_{<t}\dt\).

As in Fisher-based merge methods, we approximate this curvature with a
stored diagonal Fisher summary from previous tasks. Let
\(\Fbar_{<t}\) denote the aggregate diagonal Fisher information
computed from previous tasks before their data are discarded. We define
the Fisher-weighted interference proxy
\begin{equation}
    \qt = \langle \dt,\Fbar_{<t}\dt\rangle .
    \label{eq:fisher-interference-main}
\end{equation}
A smaller \(\qt\) indicates that the new task vector moves less in
directions estimated to be important for previous tasks, and should
therefore induce a flatter, less coefficient-sensitive write-in path.

Crucially, \(\qt\) is quadratic in \(\dt\), so the task-vector radius
directly controls a norm-based upper bound on the Fisher-weighted
curvature term. For positive semidefinite \(\Fbar_{<t}\),
\begin{equation}
    \qt
    \le
    \|\Fbar_{<t}\|_{\mathrm{op}} \|\dt\|_2^2 .
    \label{eq:fisher-norm-bound}
\end{equation}

Thus, reducing the task-vector radius lowers an upper
bound on Fisher-weighted interference. This provides a merge-rule-agnostic
way to shape mergeability: lowering this bound reduces the
curvature-sensitive variation along the write-in path
\(\theta_{t-1}+\alpha\dt\), and is expected to broaden the near-optimal
coefficient plateau \(\Wt{\varepsilon}\), provided that current-task
plasticity is not over-regularized.

This observation motivates a Fisher-free norm surrogate rather than
directly minimizing \(\qt\) during task-vector training. Directly
optimizing \(\qt\) would require additional curvature estimation and
would tie the training objective to a particular Fisher machinery,
whereas controlling the radius of \(\dt\) is geometry-agnostic. For each
LoRA product,
\begin{equation}
    \|\dt\|_F
    =
    s_{\mathrm{LoRA}}\|B_tA_t\|_F
    \le
    s_{\mathrm{LoRA}}\|B_t\|_F\|A_t\|_F .
    \label{eq:lora-product-bound}
\end{equation}
The proximal penalty controls the distance of \(A_t,B_t\) from their
initial values, and therefore controls the factor norms up to constants
set by initialization:
\[
\|A_t\|_F \le \|A_t^{(0)}\|_F + \|A_t-A_t^{(0)}\|_F,
\qquad
\|B_t\|_F \le \|B_t^{(0)}\|_F + \|B_t-B_t^{(0)}\|_F .
\]
Thus, constraining the LoRA factors provides a practical norm-based
surrogate for limiting write-in curvature. This surrogate is not the
same as directly optimizing \(\qt\); rather, it biases task-vector
training toward smaller-radius updates, thereby lowering a norm-based
upper bound on Fisher-weighted interference while leaving the write-in
coefficient rule unchanged. The full \prm{} objective and implementation
specifics are deferred to Appendix~\ref{app:implementation}.

\subsection{\prm{}: Proximal Shaping for Mergeable Task Vectors}
\label{sec:method:prm}

\prm{} instantiates the above idea by adding a proximal penalty to the
task-vector training loss. In our main instantiation, this loss uses the
same perturbation-aware forward pass as the corresponding perturbation
baseline. For current-task data
\((x,y)\sim\mathcal{D}_t\) and perturbation
\(\tilde{\epsilon}\sim\mathcal{P}_{\epsilon}\), we optimize
\begin{equation}
\begin{aligned}
\mathcal{L}_{\prm{}}
=
&
\mathbb{E}_{(x,y)\sim\mathcal{D}_t,\,
\tilde{\epsilon}\sim\mathcal{P}_{\epsilon}}
\,
\mathcal{L}_{\mathrm{CE}}
\!\left(
f_{\theta_{t-1}+(1+\tilde{\epsilon})\dt(A_t,B_t)}(x),
y
\right)
\\
&
+
\lambda_{\mathrm{prox}}
\left(
\|A_t-A_t^{(0)}\|_F^2
+
\|B_t-B_t^{(0)}\|_F^2
\right),
\end{aligned}
\label{eq:prm-objective}
\end{equation}
where \((A_t^{(0)},B_t^{(0)})\) are the initialization values of the
task-specific LoRA factors, and \(\mathcal{P}_{\epsilon}\) is the
perturbation distribution used by the corresponding perturbation-aware
baseline. The proximal penalty acts only on the current task's LoRA
factors; it does not regularize or move the running shared model
\(\theta_{t-1}\) during task-vector training.

The cross-entropy term preserves current-task plasticity, while the
proximal term limits the radius of the LoRA factor update and therefore
lowers a norm-based upper bound on \(\qt\). By reducing the
curvature-sensitive variation along \(\theta_{t-1}+\alpha\dt\), this
biases the resulting product \(\dt(A_t,B_t)\) toward a broader and less
coefficient-sensitive write-in path. This is the key distinction from
coefficient-only methods: \prm{} changes the task vector that will be
written in, not the definition of the write-in operation.

Because \prm{} is coefficient-rule agnostic, we evaluate two
instantiations:
\begin{itemize}
\setlength{\itemsep}{1pt}
\item \textbf{\prm{}-fixed}: uses a constant coefficient
\(\at=\alpha_{\mathrm{fix}}\) for all tasks. This is our default
variant, because it removes per-task Fisher coefficient estimation and
directly tests whether proximal task-vector shaping can make a single
coefficient competitive.
\item \textbf{\prm{}-Fisher}: uses the Fisher-derived write-in
coefficient rule. This variant keeps the downstream write-in rule fixed
and isolates the effect of changing the learned task vector.
\end{itemize}

After task-vector training, both variants perform the same write-in
operation in Eq.~\eqref{eq:sequential-writein}; they differ only in the
coefficient rule used for write-in.

\subsection{Metrics for Mergeability}
\label{sec:method:metrics}

We quantify mergeability with two complementary metrics: a path-level
plateau-width metric and a vector-level interference metric. We evaluate
the write-in path \(\theta_t(\alpha)\) on a fixed coefficient grid
\(\mathcal{A}=\{0.00,0.05,\dots,0.95\}\). Let
\(\overline{\mathrm{Acc}}_{\le t}(\alpha)\) denote validation accuracy
averaged over all tasks seen so far after writing in \(\dt\) with
coefficient \(\alpha\). These sweeps are used only for analysis and are
never used to select the deployed coefficient in benchmark comparisons.

\paragraph{Stable plateau width.}
For tolerance \(\varepsilon\), the near-optimal plateau width is
\begin{equation}
\Wt{\varepsilon}
=
\frac{1}{|\mathcal{A}|}
\left|
\left\{
\alpha\in\mathcal{A}:
\overline{\mathrm{Acc}}_{\le t}(\alpha)
\ge
\max_{\alpha'\in\mathcal{A}}
\overline{\mathrm{Acc}}_{\le t}(\alpha')
-
\varepsilon
\right\}
\right|.
\label{eq:plateau-width}
\end{equation}
Larger \(\Wt{\varepsilon}\) means more coefficients achieve near-optimal
seen-task accuracy. Compared with reporting only the best point on the
sweep, \(\Wt{\varepsilon}\) directly captures the property that motivates
mergeability: the write-in path is broadly usable rather than narrowly
peaked.

\paragraph{Vector-level interference.}
At the vector level, we report the Fisher-weighted interference
\(\qt\) from Eq.~\eqref{eq:fisher-interference-main} together with its
direction-normalized counterpart
\begin{equation}
    q_t^{\mathrm{dir}}
    =
    \frac{\qt}{\|\dt\|_F^2+\varepsilon_{\mathrm{num}}}.
    \label{eq:qdir}
\end{equation}
Equivalently,
\[
    \qt =
    \left(\|\dt\|_F^2+\varepsilon_{\mathrm{num}}\right)
    q_t^{\mathrm{dir}},
\]
and for non-negligible \(\|\dt\|_F\), this is approximately
\(\qt \approx \|\dt\|_F^2 q_t^{\mathrm{dir}}\). This decomposition
separates two effects: whether proximal shaping
reduces interference by shrinking the update magnitude
(\(\|\dt\|_F^2\) drops), or by reorienting the update away from
Fisher-important subspaces (\(q_t^{\mathrm{dir}}\) drops). The mechanism
analyses in \S\ref{sec:res:mechanism} use this decomposition to attribute
the gain.

All mergeability metrics are diagnostic. P\&M and \prm{}-Fisher use
the Fisher-derived write-in rule, while \prm{}-fixed uses a fixed
coefficient chosen in advance.
\section{Results}
\label{sec:results}

To answer the mergeability question posed in \S\ref{sec:intro}, we
evaluate \prm{} at three levels: benchmark performance, transfer across
settings and merge rules, and path-level attribution. The benchmarks test
whether learning more mergeable task vectors makes a fixed write-in
coefficient competitive across datasets and task splits. The paired
Base$\rightarrow$+Prox comparisons test whether the same task-vector change
improves several write-in rules and settings.
Path-level analyses then explain the effect: coefficient sweeps inspect
variation along $\theta_{t-1}+\alpha\delta_t$, norm-matched controls
separate direction from scale, the $\lambda_{\mathrm{prox}}$ sweep studies
the stability--plasticity trade-off, and ablations attribute the gain
across task-vector shaping, coefficient selection, and write-in.

\begin{figure*}[htbp]
\centering
\begin{subfigure}[t]{0.38\textwidth}
\centering
\includegraphics[width=\linewidth]{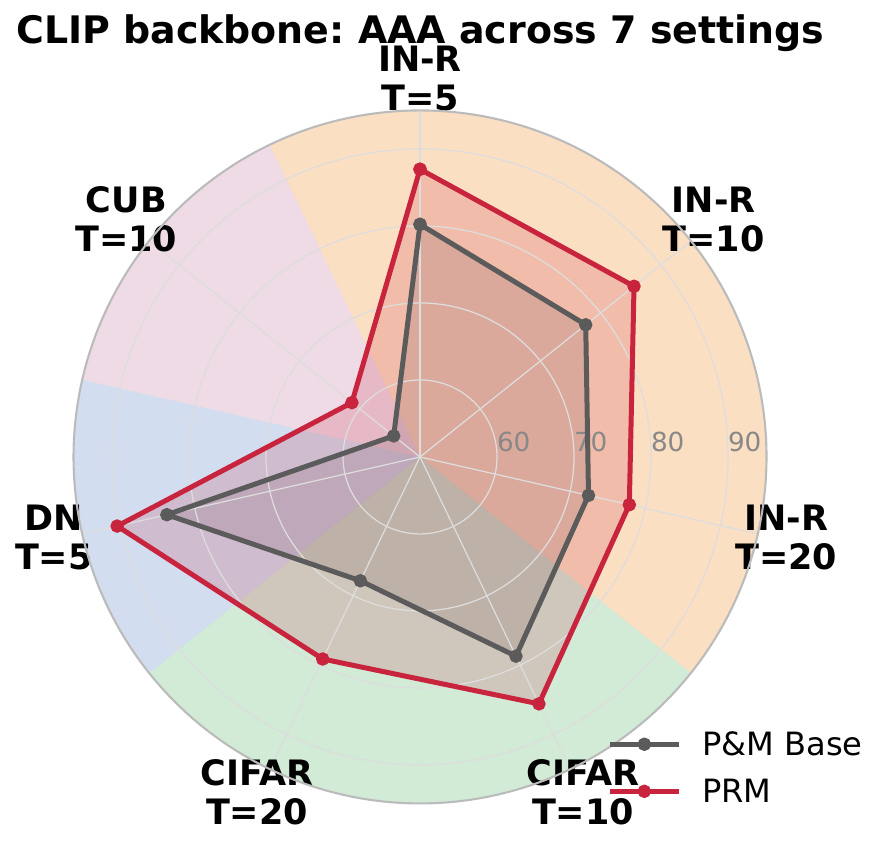}
\caption{Datasets (CLIP backbone)}
\label{fig:univ:datasets}
\end{subfigure}\hfill
\begin{subfigure}[t]{0.38\textwidth}
\centering
\includegraphics[width=\linewidth]{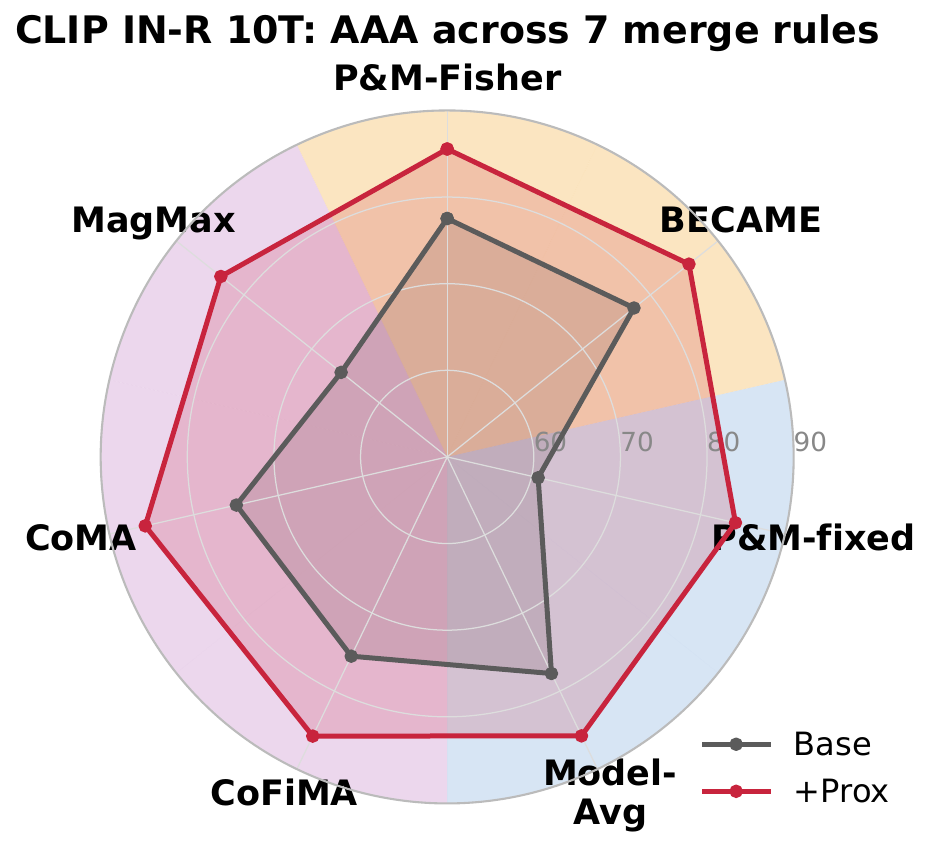}
\caption{Merge rules (CLIP IN-R 10T)}
\label{fig:univ:rules}
\end{subfigure}
\caption{Paired transfer checks for proximal task-vector shaping.
\textbf{(a)}~Dataset-level comparison on the CLIP backbone: \prm{}
improves over the corresponding P\&M Base across the plotted settings.
\textbf{(b)}~Merge-rule comparison on CLIP IN-R 10T: the same proximal
task-vector-training term improves each reported Base$\rightarrow$+Prox pair while leaving the write-in rule unchanged. Full values and setup checks
are reported in Appendix~\ref{app:beyond-pm} and Appendix~\ref{app:backbones}.}
\label{fig:universality}
\end{figure*}

\subsection{Experimental setup}
\label{sec:res:setup}

We evaluate rehearsal-free class-incremental learning on ImageNet-R~\citep{hendrycks2021many}
with 5/10/20 tasks, CIFAR-100~\citep{krizhevsky2009learning} with 10/20 tasks, DomainNet~\citep{peng2019moment} with 5 tasks,
and CUB-200~\citep{wah2011caltech} with 10 tasks. The primary metric is average anytime
accuracy (\textsc{AAA}), the mean seen-class accuracy averaged over task
boundaries. We also report final accuracy (Acc) and final forgetting. All numbers are mean$\pm$std
over three seeds unless otherwise noted.

All benchmark experiments use ViT-B/16~\citep{dosovitskiy2021image} with rank-10 LoRA adapters on
attention key/value projections. The main comparison table uses AugReg checkpoint~\citep{steiner2021train},
where baseline coverage is broadest; the paired transfer and path
diagnostics use CLIP~\citep{radford2021learning} as a diagnostic setting, where the separation
between unshaped and proximally shaped updates is clearest. Additional
backbone checks and optimization details are reported in
Appendix~\ref{app:backbones} and Appendix~\ref{app:implementation}.
\prm{} uses a single $\lambda_{\mathrm{prox}}=10^{-2}$ in all benchmark
tables and does not tune it per dataset or split.

We compare with sequential parameter-efficient fine-tuning methods
(LoRA, LoRA-EWC, Inf-LoRA, SD-LoRA), prompt-based class-incremental
methods (L2P and CODA-Prompt where available), and task-vector merging
baselines including Model-Avg, MagMax, CoFiMA, CoMA, BECAME, and P\&M.
We report two variants: \prm{}-fixed uses a constant coefficient
$\alpha{=}0.8$, while \prm{}-Fisher keeps P\&M's Fisher-derived rule to
isolate the effect of proximal task-vector shaping. Unless specified,
\prm{} denotes \prm{}-fixed.

\begin{table*}[htbp]
\centering
\caption{Results on ImageNet-R (ViT-B/16 AugReg). \textbf{\prm{}-fixed} and \textbf{\prm{}-Fisher} are ours.}
\label{tab:main-inr}
\setlength{\tabcolsep}{3pt}
\renewcommand{\arraystretch}{1.05}
\resizebox{\textwidth}{!}{%
\begin{tabular}{ccccccc}
\toprule
 & \multicolumn{6}{c}{ImageNet-R} \\
\cmidrule(lr){2-7}
 & \multicolumn{2}{c}{T=5} & \multicolumn{2}{c}{T=10} & \multicolumn{2}{c}{T=20} \\
\cmidrule(lr){2-3} \cmidrule(lr){4-5} \cmidrule(lr){6-7}
Method & Acc\,$\uparrow$ & AAA\,$\uparrow$ & Acc\,$\uparrow$ & AAA\,$\uparrow$ & Acc\,$\uparrow$ & AAA\,$\uparrow$ \\
\midrule
Full-FT & 64.92\,$\pm$\,.87$^{\dagger}$ & 75.57\,$\pm$\,.50$^{\dagger}$ & 60.57\,$\pm$\,1.06$^{\dagger}$ & 72.31\,$\pm$\,1.09$^{\dagger}$ & 49.95\,$\pm$\,1.31$^{\dagger}$ & 65.32\,$\pm$\,.69$^{\dagger}$ \\
L2P & 73.04\,$\pm$\,.71$^{\dagger}$ & 76.94\,$\pm$\,.61$^{\dagger}$ & 71.26\,$\pm$\,.44$^{\dagger}$ & 76.13\,$\pm$\,.46$^{\dagger}$ & 68.97\,$\pm$\,.51$^{\dagger}$ & 74.16\,$\pm$\,.32$^{\dagger}$ \\
BECAME & 81.69\,$\pm$\,.25 & 86.25\,$\pm$\,.37 & 80.05\,$\pm$\,.22 & 85.65\,$\pm$\,.80 & 76.60\,$\pm$\,.45 & 83.32\,$\pm$\,.52 \\
CoFiMA & 81.99\,$\pm$\,.25 & \underline{86.37\,$\pm$\,.47} & 80.13\,$\pm$\,.37 & 85.80\,$\pm$\,.60 & 76.28\,$\pm$\,.41 & \underline{83.58\,$\pm$\,.40} \\
CoMA & 81.69\,$\pm$\,.17 & 86.21\,$\pm$\,.51 & 79.88\,$\pm$\,.61 & 85.80\,$\pm$\,.51 & 75.77\,$\pm$\,.36 & 83.37\,$\pm$\,.53 \\
MagMax & 79.39\,$\pm$\,.13 & 82.93\,$\pm$\,.56 & 79.11\,$\pm$\,.25 & 83.43\,$\pm$\,.65 & 77.07\,$\pm$\,.40 & 82.69\,$\pm$\,.63 \\
Model-Avg & 81.90\,$\pm$\,.34 & 86.32\,$\pm$\,.53 & 80.22\,$\pm$\,.21 & 85.81\,$\pm$\,.84 & 76.66\,$\pm$\,.41 & 83.43\,$\pm$\,.49 \\
EWC-LoRA & 81.19\,$\pm$\,.39 & 84.85\,$\pm$\,.44 & 78.08\,$\pm$\,.27 & 82.81\,$\pm$\,1.00 & 72.05\,$\pm$\,.87 & 77.45\,$\pm$\,1.19 \\
Inf-LoRA & 74.77\,$\pm$\,.25$^{\dagger}$ & 78.18\,$\pm$\,.24$^{\dagger}$ & 74.65\,$\pm$\,.14$^{\dagger}$ & 78.15\,$\pm$\,.24$^{\dagger}$ & 73.59\,$\pm$\,.19$^{\dagger}$ & 77.93\,$\pm$\,.19$^{\dagger}$ \\
SD-LoRA & 79.15\,$\pm$\,.20$^{\dagger}$ & 83.01\,$\pm$\,.42$^{\dagger}$ & 77.34\,$\pm$\,.35$^{\dagger}$ & 82.04\,$\pm$\,.24$^{\dagger}$ & 75.26\,$\pm$\,.37$^{\dagger}$ & 80.22\,$\pm$\,.72$^{\dagger}$ \\
LoRA & 72.96\,$\pm$\,.91 & 79.39\,$\pm$\,.57 & 66.54\,$\pm$\,1.19 & 76.72\,$\pm$\,.81 & 56.92\,$\pm$\,1.17 & 72.24\,$\pm$\,.58 \\
P\&M & \underline{82.00\,$\pm$\,.30} & \textbf{86.54\,$\pm$\,.43} & 79.96\,$\pm$\,.51 & 85.55\,$\pm$\,.90 & 76.59\,$\pm$\,.47 & 83.17\,$\pm$\,.40 \\
\midrule
\textbf{\prm{}-fixed} & \textbf{82.05\,$\pm$\,.39} & 86.03\,$\pm$\,.52 & \textbf{81.32\,$\pm$\,.25} & \textbf{86.39\,$\pm$\,.53} & \textbf{78.12\,$\pm$\,.57} & \textbf{83.70\,$\pm$\,.81} \\
\textbf{\prm{}-Fisher} & 81.64\,$\pm$\,.43 & 85.78\,$\pm$\,.42 & \underline{80.87\,$\pm$\,.44} & \underline{85.96\,$\pm$\,.48} & \underline{77.93\,$\pm$\,.30} & 83.14\,$\pm$\,.92 \\
\bottomrule
\end{tabular}%
}
\\[3pt]
{\footnotesize Each cell is mean$\pm$std over 3 seeds; $\dagger$=lifted from \citep{qiu2025pm} Tabs.~2/4/5/9/11. \textbf{Best per column bold}; \underline{second best underlined}.}
\end{table*}

\subsection{Benchmark consequence: a fixed coefficient becomes competitive}
\label{sec:res:main}

Table~\ref{tab:main-inr} reports the ImageNet-R comparison on the
AugReg backbone. If the learned task vector is stable over a range of
write-in coefficients, a single fixed coefficient should not be heavily
penalized relative to a Fisher-derived coefficient chosen at each task.

On ImageNet-R, \prm{}-fixed achieves the best final accuracy in all
three task partitions and the best \textsc{AAA} for T${=}10$ and
T${=}20$. It does not dominate every column: at T${=}5$, P\&M has higher
\textsc{AAA} by $0.51$ points. Thus, the main result is not pointwise
dominance, but that a fixed coefficient remains competitive with
Fisher-based P\&M and improves the harder ImageNet-R partitions.

Full AugReg results on CIFAR-100, DomainNet, and CUB-200 are reported in
Appendix~\ref{app:per-dataset}. Across the seven reported AugReg
settings, \prm{}-fixed obtains the highest mean \textsc{AAA} ($86.87$),
compared with $86.38$ for P\&M, while using the same $\alpha{=}0.8$ on
every task. \prm{}-Fisher also improves the mean \textsc{AAA} to
$86.75$ while keeping P\&M's Fisher-derived write-in rule. The two
variants therefore show that, once task vectors are proximally shaped,
a fixed rule can be competitive in practice, and that the gain is not
explained solely by changing the coefficient rule.

\subsection{Proximal shaping transfers across settings and write-in rules}
\label{sec:res:universality}

To test whether proximal shaping transfers beyond P\&M, we hold each write-in
rule fixed and change only task-vector training: Base uses the original recipe,
and +Prox adds the same $\lambda_{\mathrm{prox}}=10^{-2}$ penalty. We run this
diagnostic on CLIP; per-cell values and backbone checks are in
Appendix~\ref{app:beyond-pm} and Appendix~\ref{app:backbones}.

Figure~\ref{fig:universality}(a) shows positive P\&M Base$\rightarrow$\prm{}
gains across the plotted CLIP settings. Figure~\ref{fig:universality}(b)
changes the write-in rule while keeping the proximal modification fixed;
every reported Base$\rightarrow$+Prox comparison improves on CLIP IN-R
10T, and the +Prox variants occupy a narrower \textsc{AAA} range than
their Base counterparts. These paired comparisons locate the gain at
task-vector shaping. They do not, however, show how the resulting task vector behaves
as the write-in coefficient varies, so we next inspect the path
$\theta_{t-1}+\alpha\delta_t$ directly.

\subsection{Write-in paths under coefficient sweeps}
\label{sec:res:mechanism}

For a trained task vector $\delta_t$, we evaluate
$\theta_t(\alpha)=\theta_{t-1}+\alpha\delta_t$ over a fixed coefficient
grid and measure seen-task loss and accuracy. This checks the
mergeability property defined in \S\ref{sec:method:metrics}: whether
many coefficients give similar seen-task performance.

\begin{figure*}[!t]
\centering
\includegraphics[width=0.83\textwidth]{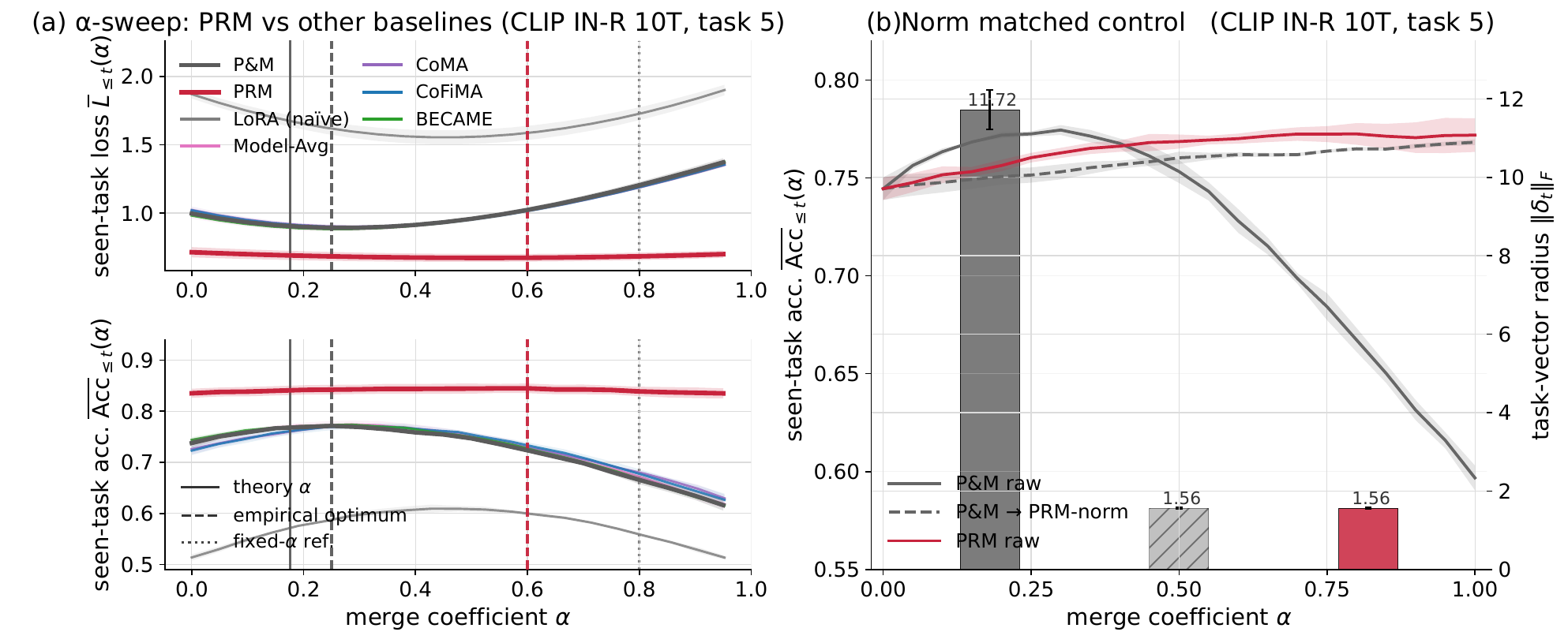}
\caption{Seen-task loss and accuracy along the write-in coefficient on CLIP
ViT-B/16 IN-R 10T at task~5. \textbf{(a)}~\prm{} has the lowest loss, highest
accuracy, and smallest variation; the fixed $\alpha=0.8$ lies near its empirical
best. \textbf{(b)}~Rescaling the P\&M vector to the \prm{} norm greatly reduces
accuracy variation, indicating that scale explains much of the coefficient
sensitivity. Bars show Frobenius norms; curves show seen-task accuracy.}
\label{fig:alpha-sweeps}
\end{figure*}

Figure~\ref{fig:alpha-sweeps}(a) reports a representative sweep at task~5
on CLIP ViT-B/16 IN-R 10T. Across the plotted coefficient grid, LoRA gives the largest loss and lowest accuracy, the write-in
baselines occupy an intermediate regime, and \prm{} gives the lowest loss
and highest accuracy with the smallest variation. The fixed reference
$\alpha{=}0.8$ therefore achieves accuracy close to the empirical maximum
of the \prm{} path without searching over $\alpha$ for that task. In this
representative sweep, \prm{} is not merely selecting a different point on
the same brittle path; proximal shaping changes the path induced by the
task vector.

\paragraph{Norm-matched control.}
The reduced variation in Figure~\ref{fig:alpha-sweeps}(a) could come
from a changed direction, or from traversing a shorter distance because
$\|\delta_t\|_F$ is smaller. We use P\&M because it is the closest
unproximal counterpart to \prm{}: it shares the perturbation-aware
task-vector training but lacks the proximal penalty. To separate
direction from scale, we preserve the P\&M direction and replace only
its Frobenius norm with the \prm{} norm. Figure~\ref{fig:alpha-sweeps}(b)
shows that rescaling the P\&M vector from norm $11.72$ to $1.56$
substantially reduces accuracy variation at large $\alpha$. The rescaled
curve does not coincide with raw \prm{}, so directional or other
task-vector-training effects may still contribute, but scale explains a
large part of the coefficient sensitivity in this case.
Appendix~\ref{app:causal} reports the corresponding continuous scaling
control.

\subsection{Proximal strength and the stability--plasticity trade-off}
\label{sec:res:lambda}

The norm-matched and scaling controls identify task-vector scale as an
important source of coefficient sensitivity, but they do not determine
how strongly task-vector training should be regularized. We therefore sweep
$\lambda_{\mathrm{prox}}$ on CLIP IN-R 10T and measure both diagnostic
path metrics and full benchmark performance; the exact grid is reported
in Appendix~\ref{app:implementation}.

Figure~\ref{fig:lambda-sensitivity}(a) reports the diagnostic metrics
from \S\ref{sec:method:metrics}. $W_{t,\varepsilon}$ measures the
fraction of coefficients within $\varepsilon$ of the best sweep point;
$q_t$ is Fisher-weighted interference; and
$q_t^{\mathrm{dir}}=q_t/(\|\delta_t\|_F^2+\varepsilon_{\mathrm{num}})$
separates update scale from direction. As $\lambda_{\mathrm{prox}}$
increases, $W_{t,\varepsilon}$ increases and $q_t$ drops by orders of
magnitude, whereas $q_t^{\mathrm{dir}}$ is not monotonic. Since
$q_t\approx\|\delta_t\|_F^2 q_t^{\mathrm{dir}}$, the reduction in
interference is mainly explained by a smaller update norm rather than a
monotonic rotation away from Fisher-important subspaces.

Figure~\ref{fig:lambda-sensitivity}(b) shows the corresponding
benchmark trade-off. \textsc{AAA} and Acc improve from the
no-proximal setting to intermediate values, are best around
$\lambda_{\mathrm{prox}}\in[3{\times}10^{-3},10^{-2}]$, and decrease
when the proximal penalty becomes too strong. Thus, reducing the
task-vector norm helps only while current-task plasticity is preserved.
We use $\lambda_{\mathrm{prox}}{=}10^{-2}$ because it lies in the
high-performing range before the large-$\lambda_{\mathrm{prox}}$
degradation.

\begin{figure*}[htbp]
\centering
\includegraphics[width=0.83\textwidth]{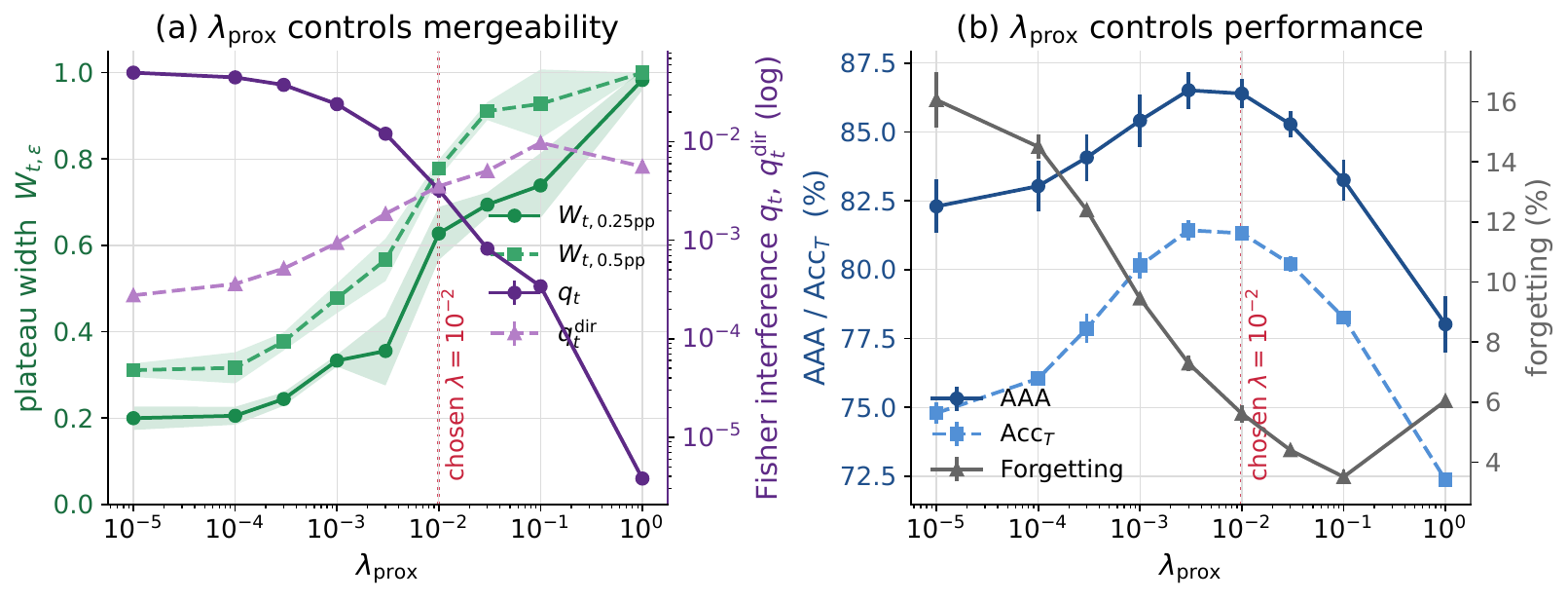}
\caption{Effect of proximal strength on path diagnostics and benchmark
performance. The $x$-axis is logarithmic and $\lambda_{\mathrm{prox}}=0$
is mapped to $10^{-5}$ for visualization. \textbf{(a)}~As
$\lambda_{\mathrm{prox}}$ increases, the fraction of coefficients within
0.25 and 0.5 percentage points of the sweep optimum increases, while
$q_t$ decreases by orders of magnitude. The direction-normalized
interference $q_t^{\mathrm{dir}}$ is not monotonic. \textbf{(b)}~Benchmark
performance is highest at an intermediate proximal strength and decreases
for very large $\lambda_{\mathrm{prox}}$.}
\label{fig:lambda-sensitivity}
\end{figure*}

\subsection{Attributing the gain across task-vector and write-in components}
\label{sec:res:ablation}

Table~\ref{tab:ablation} separates perturbation-aware training, proximal
shaping, the coefficient rule, and the final write-in. The lower NoWrite block
sets $\alpha=0$ for $t\geq1$ and is used only for attribution.

\begin{table*}[htbp]
\centering
\caption{Ablation and NoWrite attribution on ImageNet-R 10T (3 seeds,
mean$\pm$std). $\overline{W}_{t,0.01}$ is near-optimal plateau width and
$\overline{q}_t$ is Fisher-weighted interference in $10^{-3}$ units, averaged
over sweep tasks $\{2,5,10\}$.}
\label{tab:ablation}
\setlength{\tabcolsep}{3pt}
\renewcommand{\arraystretch}{1.05}
\resizebox{\textwidth}{!}{%
\begin{tabular}{lccc ccc cc}
\toprule
Method & pert & prox & $\alpha$-rule
& \textsc{AAA} & \textsc{Acc}$_T$ & \textsc{Forg}
& $\overline{W}_{t,0.01}$ & $\overline{q}_t$ [$10^{-3}$] \\
\midrule
\multicolumn{9}{l}{\emph{Upper block: deployable variants}} \\
LoRA baseline
& -- & -- & $\alpha{=}1$
& $76.72{\pm}0.81$ & $66.54{\pm}1.19$ & $26.66{\pm}1.73$
& 0.38 & 63.94 \\
P\&M-fixed
& \checkmark & -- & $\alpha{=}0.8$
& $82.30{\pm}0.97$ & $74.80{\pm}0.40$ & $16.06{\pm}0.94$
& $0.44$ & $45.12$ \\
P\&M
& \checkmark & -- & Fisher
& $85.55{\pm}0.90$ & $79.96{\pm}0.51$ & $3.50{\pm}0.41$
& $0.44$ & $45.12$ \\
LoRA-Prox
& -- & \checkmark & $\alpha{=}0.8$
& $85.61{\pm}0.48$ & $80.27{\pm}0.31$ & $7.72{\pm}0.62$
& 0.84 & $4.93$ \\
\prm{}-fixed
& \checkmark & \checkmark & $\alpha{=}0.8$
& $86.39{\pm}0.53$ & $81.32{\pm}0.25$ & $5.62{\pm}0.29$
& $0.91$ & $3.26$ \\
\prm{}-Fisher
& \checkmark & \checkmark & Fisher
& $85.96{\pm}0.48$ & $80.87{\pm}0.44$ & $4.28{\pm}0.28$
& $0.91$ & $3.56$ \\
\midrule
\multicolumn{9}{l}{\emph{Lower block: NoWrite (attribution-only; $\alpha{=}0$ for $t\geq1$)}} \\
LoRA-NoWrite
& -- & -- & $\alpha{=}0$
& $72.88{\pm}1.42$ & $65.73{\pm}1.12$ & $3.05{\pm}0.36$
& 0.38 & 63.94 \\
P\&M-NoWrite
& \checkmark & -- & $\alpha{=}0$
& $76.64{\pm}0.74$ & $69.97{\pm}0.98$ & $3.10{\pm}0.44$
& $0.44$ & $45.12$ \\
LoRA-Prox-NoWrite
& -- & \checkmark & $\alpha{=}0$
& $82.78{\pm}0.33$ & $76.95{\pm}0.20$ & $5.00{\pm}0.31$
& 0.84 & $5.62$ \\
\prm{}-NoWrite
& \checkmark & \checkmark & $\alpha{=}0$
& $84.45{\pm}0.33$ & $78.77{\pm}0.07$ & $4.98{\pm}0.12$
& $0.91$ & $3.26$ \\
\bottomrule
\end{tabular}%
}
\end{table*}

Merely merging the original P\&M
vector with $\alpha{=}0.8$ gives \textsc{AAA}$=82.30\pm0.97$; the
Fisher-derived P\&M rule recovers to $85.55\pm0.90$, while
\prm{}-fixed reaches $86.39\pm0.53$ with the same fixed coefficient.
The path diagnostics move in the same direction: \prm{}-fixed increases
$\overline{W}_{t,0.01}$ from $0.44$ to $0.91$ and reduces
$\overline{q}_t$ from $45.12$ to $3.26$ in $10^{-3}$ units.

The next comparisons localize the gain within the pipeline. LoRA-Prox,
which adds the proximal term without P\&M's perturbation-aware objective,
already reaches \textsc{AAA}$=85.61\pm0.48$ and reduces
$\overline{q}_t$ to $4.93\times10^{-3}$, suggesting that proximal shaping
is the component most directly associated with the mergeability change in
this ablation; adding perturbation-aware training improves further to
$86.39\pm0.53$. Once the proximal term is
present, the coefficient rule has limited remaining effect in this setting:
\prm{}-fixed and \prm{}-Fisher differ by only $0.43$ \textsc{AAA}
points and have nearly identical diagnostics
($\overline{W}_{t,0.01}=0.91$, $\overline{q}_t=3.26$ versus $3.56$ in
$10^{-3}$ units).

The NoWrite rows isolate performance prior to write-in. The
ordering is LoRA-NoWrite ($72.88$) $<$ P\&M-NoWrite ($76.64$)
$<$ LoRA-Prox-NoWrite ($82.78$) $<$ \prm{}-NoWrite ($84.45$)
$<$ full \prm{}-fixed ($86.39$). The final write-in step still adds
$1.94$ \textsc{AAA} points over \prm{}-NoWrite, but \prm{}-NoWrite is
only $1.10$ points below full P\&M. Thus, much of the improvement is
already present in the shaped task vector before it is written into
the running model. Overall, the benchmark gains trace primarily to
proximal task-vector shaping rather than to a new write-in coefficient
rule or to a trivial collapse of the update.


\section{Conclusion and Limitations}

We propose PRM, a proximal regularization method for rehearsal-free continual LoRA merging. It shapes task-vector training by penalizing current LoRA factors' distance from initialization, avoiding new merge-coefficient rules. Across multiple experiment settings, PRM consistently
improves its base variants and matches or outperforms strong coefficient-based
baselines. Diagnostics show that proximal shaping broadens the usable range of
merge coefficients, reduces Fisher-weighted interference, and moves task
vectors into a smaller-radius write-in regime. These results suggest that
effective sequential LoRA merging depends not only on how a task vector is
scaled at merge time, but also on whether it has been trained to induce a stable
write-in path.

\textbf{Limitations.}
Our empirical scope remains limited to rehearsal-free class-incremental vision
benchmarks, leaving broader continual-learning regimes, modalities, and longer
streams for future study. PRM is also studied mainly under additive LoRA
write-in, and its extension to other adapters, full-model updates, or
non-additive merging rules remains open. Future work should evaluate PRM in broader settings, develop richer magnitude- and direction-aware task-vector constraints, and
further clarify its connection to the local loss landscape.

\bibliographystyle{plainnat}
\bibliography{refs}

\clearpage
\appendix

\section*{Appendix}

\section{Experimental details and reproducibility}
\label{app:details}
\label{app:implementation}  

This appendix collects the experimental protocol details referenced in
the main text.  We keep the description at the level needed for
reproducibility, and avoid implementation-specific names.

\paragraph{A.0 Code availability.}
The initial anonymous submission does not include a code release. The method,
training objective, experimental protocol, hyperparameters, baseline implementation
details, diagnostic metrics, and compute resources are described in this appendix to
support reproducibility. We plan to release the code and configuration files upon
acceptance.

\paragraph{A.1 Task construction and evaluation.}
We evaluate rehearsal-free class-incremental learning on four image
datasets and seven task splits: ImageNet-R with $5$, $10$, and $20$
tasks; CIFAR-100 with $10$ and $20$ tasks; DomainNet with $5$ tasks;
and CUB-200 with $10$ tasks.  For each (dataset, $T$) pair we use a
fixed class permutation that is shared across all in-house methods so
that comparisons are paired at the seed level; previous-task data are
not available during task-vector training for any subsequent task.  Lifted
baselines marked with $\dagger$ in the result tables are reused under
the matching protocol of \citet{qiu2025pm}.

Let $a_{t,j}\in[0,1]$ denote the validation accuracy on task $j$
measured after training through task $t$ (with $a_{t,j}\!=\!0$ by
convention for $t<j$).  We report average anytime accuracy
(\textsc{AAA}), final accuracy (\textsc{Acc}$_T$), and final
forgetting:
\begin{align}
\mathrm{AAA}
&=
\frac{1}{T}\sum_{t=1}^{T}
\frac{1}{t}\sum_{j=1}^{t} a_{t,j},
&
\mathrm{Acc}_T
&=
\frac{1}{T}\sum_{j=1}^{T}a_{T,j},
\notag\\
\mathrm{Forg}
&=
\frac{1}{T-1}\sum_{j=1}^{T-1}
\Bigl(
\max_{t\in\{j,\ldots,T\}} a_{t,j}
-
a_{T,j}
\Bigr).
\label{eq:metrics}
\end{align}
The maximum in the forgetting expression is taken over all training
times that have observed task $j$, including the final state
$t{=}T$, which makes the per-task forgetting non-negative.  This
matches the non-negative variant of \citet{chaudhry2018rwalk} and is
the metric used in our codebase.  All tables in the main text and
appendix report mean$\pm$std over three seeds unless explicitly
marked otherwise.

\paragraph{A.2 Backbone, LoRA, head, and optimizer.}
The vision backbone is a ViT-B/16~\citep{dosovitskiy2021image} frozen during continual training.
We use the AugReg checkpoint~\citep{steiner2021train} 
\texttt{vit\_base\_patch16\_224.augreg2\_in21k\_ft\_in1k} for the main
benchmark study and the CLIP-OpenAI checkpoint~\citep{radford2021learning} 
\texttt{vit\_base\_patch16\_clip\_224.openai} for the CLIP transfer
study; the two backbones share the same training pipeline and differ
only in the checkpoint and the cosine-head temperature ($30$ for
AugReg, $28$ for CLIP-OpenAI).  Each adapted attention layer carries
rank-10 LoRA factors $(A_t,B_t)$ on the key and value projections,
with $A_t$ initialised by Kaiming-uniform (gain $\sqrt{5}$) and $B_t$
initialised to zero so that the new task contributes the zero
write-in at $t{=}0$ epochs.  We apply no additional LoRA scaling,
i.e.\ the effective task vector is $\dt=B_t A_t$; expressed in the
notation of Eq.~(\ref{eq:effective-task-vector}) this corresponds to
$s_{\mathrm{LoRA}}{=}1$.  Only the per-task LoRA factors and the
per-task linear classifier head are updated during task-vector training.
Task-vector training uses AdamW~\citep{loshchilov2019decoupled} with base learning rate $10^{-3}$, classifier-head
learning-rate multiplier $10$ (so the head trains at $10^{-2}$), batch
size $256$, and a multi-step learning-rate schedule with no warmup;
the default schedule runs for $10$ epochs per task on every dataset
except DomainNet, which uses $5$ epochs per task.  When the P\&M
perturbation objective is used, the perturbation magnitude is fixed at
$\epsilon{=}0.5$ with sampling probability $p_0{=}0.33$, following
\citet{qiu2025pm}; the perturbed forward pass is then
$f_{\theta_{t-1}+(1+\tilde{\epsilon})\dt}(x)$.

\paragraph{A.3 \prm{} objective and algorithm.}
For current-task data $(x,y)\sim\mathcal{D}_t$ and perturbation
$\tilde{\epsilon}\sim\mathcal{P}_{\epsilon}$, \prm{}
optimizes the following task-vector-training objective:
\[
\mathcal{L}_{\prm{}}
=
\mathbb{E}_{(x,y)\sim\mathcal{D}_t,\,
\tilde{\epsilon}\sim\mathcal{P}_{\epsilon}}
\mathcal{L}_{\mathrm{CE}}\!\left(
 f_{\theta_{t-1}+(1+\tilde{\epsilon})\dt(A_t,B_t)}(x),y
\right)
+
\lambda_{\mathrm{prox}}
\left(
\|A_t-A_t^{(0)}\|_F^2 + \|B_t-B_t^{(0)}\|_F^2
\right),
\]
where $(A_t^{(0)},B_t^{(0)})$ are the LoRA-factor initializations.
The proximal term acts only on the current task's LoRA factors;
$\theta_{t-1}$ is not regularized during task-vector training.  We use
$\lambda_{\mathrm{prox}}{=}10^{-2}$ in all main-table experiments;
the proximal-strength sweep grid for the diagnostic in
\S\ref{sec:res:lambda} is listed in paragraph~A.5.
Algorithm~\ref{alg:prm} summarizes the per-task pipeline.

\begin{algorithm}[t]
\caption{\prm{} for rehearsal-free sequential LoRA merging.}
\label{alg:prm}
\begin{algorithmic}
\REQUIRE Pretrained backbone $\theta_0$; task stream
  $\{\mathcal{D}_t\}_{t=1}^{T}$; proximal weight $\lambda_{\mathrm{prox}}$;
  coefficient rule (\prm{}-fixed: $\at{=}0.8$; \prm{}-Fisher: the
  Fisher rule below).
\STATE Initialise running model with $\theta_{0}$ (and an empty
  Fisher summary $\Fbar_{0}\!\leftarrow\!0$ if the coefficient rule
  uses Fisher information).
\FOR{$t=1,\dots,T$}
\STATE \textbf{Step 1 (initialise task $t$).}
  Freeze $\theta_{t-1}$.  For every adapted attention layer
  $\ell$, allocate a fresh LoRA pair $(A_t^{(0)},B_t^{(0)})$ with
  $A_t^{(0)}\!\sim\!\mathrm{Kaiming\text{-}Uniform}(\sqrt{5})$ and
  $B_t^{(0)}\!=\!0$ so that the initial task vector
  $\dt^{(0)}{=}B_t^{(0)}A_t^{(0)}{=}0$ and the model is unchanged
  before training.  Allocate a fresh per-task linear classifier head.
\STATE \textbf{Step 2 (train the task-vector proposal).}
  Optimise $(A_t,B_t)$ and the head on $\mathcal{D}_t$ with AdamW
  for the configured number of epochs:
  \begin{itemize}
  \setlength{\itemsep}{1pt}
  \item[\hspace{1em}(2a)] At each step, sample a perturbation
    $\tilde{\epsilon}\!\sim\!\mathcal{P}_{\epsilon}$ (zero with
    probability $1{-}p_0$); compute logits at
    $\theta_{t-1}+(1{+}\tilde{\epsilon})B_tA_t$ and the cross-entropy
    loss $\mathcal{L}_{\mathrm{CE}}$.
  \item[\hspace{1em}(2b)] Add the proximal regulariser
    $\lambda_{\mathrm{prox}}\bigl(\|A_t-A_t^{(0)}\|_F^{2}+\|B_t-B_t^{(0)}\|_F^{2}\bigr)$
    on the current-task LoRA factors; the running shared model
    $\theta_{t-1}$ is not regularised.
  \item[\hspace{1em}(2c)] Backpropagate and apply the AdamW update.
  \end{itemize}
  Form the effective task vector $\dt\!\leftarrow\!B_t A_t$ across
  all adapted projections.
\STATE \textbf{Step 3 (choose the write-in coefficient).}\\
  \(\bullet\) \prm{}-fixed: set $\at{=}0.8$.\\
  \(\bullet\) \prm{}-Fisher: estimate the per-task diagonal Fisher
  $\hat F_t$ on $\mathcal{D}_t$ over the K and V projection weights
  by averaging
  $\bigl(\nabla_{W_{kv}^{(\ell)}}\mathcal{L}_{\mathrm{CE}}(x,y)\bigr)^{2}$
  over the task data, and form the closed-form Fisher rule from
  \citet{qiu2025pm}:
  \[
  \alpha_t
  =
  -\,\frac{
    \displaystyle\sum_{\ell}\sum_{i=0}^{t}
      \bigl\langle
        F_i^{(\ell)}\!\odot\!\dt^{(\ell)},\;
        W_{<t}^{(\ell)}-\widetilde W_i^{(\ell)}-\dt^{(\ell)}
      \bigr\rangle
  }{
    \displaystyle\sum_{\ell}\sum_{i=0}^{t}
      \bigl\langle
        F_i^{(\ell)}\!\odot\!\dt^{(\ell)},\;
        \dt^{(\ell)}
      \bigr\rangle
  },
  \]
  where the sum runs over adapted layers $\ell$ and over previously
  observed tasks $i\!\in\!\{0,\dots,t\}$;
  $W_{<t}^{(\ell)}{=}\sum_{j<t}\alpha_j\,B_jA_j$ is the accumulated
  LoRA delta on layer $\ell$ before the current task;
  $\widetilde W_i^{(\ell)}{=}\sum_{j<i}\alpha_j\,B_jA_j$ is the
  accumulated delta evaluated at the snapshot when task $i$ was
  trained; and $\odot$ is the elementwise product on
  $\mathbb{R}^{d_{\mathrm{out}}\times d_{\mathrm{in}}}$.  The Fisher
  diagonal is sliced from the K and V row blocks of the QKV weight
  matrix.
\STATE \textbf{Step 4 (write back \& evaluate).}
  Update the running model
  $\theta_t \leftarrow \theta_{t-1} + \at\,\dt$, and (when needed by
  the coefficient rule) accumulate the diagonal Fisher summary by
  storing $\hat F_t$ alongside $\Fbar_{t-1}$.  Evaluate
  $\theta_t$ on the validation splits of all tasks $j\!\le\!t$ to
  populate the row $a_{t,\cdot}$ of the accuracy matrix used in
  Eq.~(\ref{eq:metrics}).
\ENDFOR
\end{algorithmic}
\end{algorithm}

\paragraph{A.4 Compared methods and implementation sources.}
Table~\ref{tab:baseline-implementation} groups the methods compared
in \S\ref{sec:res:main} and \S\ref{sec:res:universality} by their
task-vector-training objective and write-in rule.  All in-house runs use
the same task splits, backbone, LoRA rank, optimizer, and evaluation
protocol described in paragraphs~A.1--A.2.  Cells marked with
$\dagger$ in result tables are lifted under the matching protocol of
\citet{qiu2025pm}; cells without $\pm$ aggregate a single available
seed.  $+$Prox rows in Appendix~\ref{app:universality} keep the
listed write-in rule and only modify task-vector training by adding the proximal
penalty with the same $\lambda_{\mathrm{prox}}{=}10^{-2}$.

\begin{table}[t]
\centering
\small
\setlength{\tabcolsep}{3.5pt}
\renewcommand{\arraystretch}{1.05}
\caption{Compared methods and implementation sources.  All in-house
runs share the protocol of paragraphs~A.1--A.2.  Dagger entries in
result tables are lifted under the matching protocol of
\citet{qiu2025pm}; cells without $\pm$ aggregate a single available
seed.  ``Same task-vector training'' is shorthand for the perturbation-aware LoRA
recipe used by P\&M; ``+ proximal'' adds the penalty in
Eq.~(\ref{eq:prm-objective}).}
\label{tab:baseline-implementation}
\begin{tabular}{|p{0.16\textwidth}|p{0.27\textwidth}|p{0.26\textwidth}|p{0.22\textwidth}|}
\hline
\textbf{Method} & \textbf{Task-vector training} & \textbf{Write-in rule} & \textbf{Source} \\
\hline\hline
Full-FT & joint update of all parameters & no merge (single model) & in-house \\
\hline
LoRA & per-task LoRA, no perturbation & write-in with $\alpha{=}1$ & in-house \\
\hline
EWC-LoRA & per-task LoRA + EWC penalty & write-in with $\alpha{=}1$ & in-house \\
\hline
L2P, DualPrompt, CODA-Prompt & prompt-pool training (no LoRA merge) & no merge (prompt routing) & lifted from \citet{qiu2025pm} where dagger-marked \\
\hline
Inf-LoRA & subspace-projected LoRA training & subspace consolidation & lifted from \citet{qiu2025pm} where dagger-marked \\
\hline
SD-LoRA & soft-router LoRA training & router-weighted consolidation & lifted from \citet{qiu2025pm} where dagger-marked \\
\hline\hline
P\&M & perturbation-aware LoRA & Fisher-derived $\at$ \citep{qiu2025pm} & in-house \\
\hline
P\&M-fixed & same task-vector training as P\&M & fixed $\at{=}0.8$ & in-house \\
\hline
Model-Avg & same task-vector training as P\&M & uniform model averaging & in-house \\
\hline
CoMA & same task-vector training as P\&M & continual model averaging \citep{marouf2024weighted} & in-house \\
\hline
CoFiMA & same task-vector training as P\&M & Fisher-weighted CoMA \citep{marouf2024weighted} & in-house \\
\hline
BECAME & same task-vector training as P\&M & adaptive-coefficient merging \citep{li2025became} & in-house \\
\hline
MagMax & same task-vector training as P\&M & maximum-magnitude selection \citep{marczak2024magmax} & in-house \\
\hline\hline
\prm{}-fixed & same task-vector training + proximal penalty & fixed $\at{=}0.8$ & in-house (this work) \\
\hline
\prm{}-Fisher & same task-vector training + proximal penalty & Fisher-derived $\at$ \citep{qiu2025pm} & in-house (this work) \\
\hline
$+$Prox (rule) & task-vector training of (rule) + proximal penalty & write-in rule of (rule), unchanged & in-house (this work) \\
\hline
\end{tabular}
\end{table}

\paragraph{A.5 Diagnostic protocols and path metrics.}
Three diagnostic axes are reused throughout the paper.  The
\emph{coefficient ($\alpha$) sweep} freezes a $(\theta_{t-1},\dt)$
pair and evaluates the merged model at every
$\alpha\in\{0.00,0.05,\dots,0.95\}$ on the validation splits of all
tasks seen so far.  The \emph{$\lambda_{\mathrm{prox}}$ sweep} used in
Figure~3 of the main text uses the grid
$\{0,\,10^{-4},\,3{\cdot}10^{-4},\,10^{-3},\,3{\cdot}10^{-3},\,10^{-2},\,3{\cdot}10^{-2},\,10^{-1},\,1\}$;
path metrics from these sweeps are averaged over tasks $\{2,5,10\}$
on ImageNet-R 10T when reported.  The \emph{NoWrite probe} is a
diagnostic-only write-in rule that sets $\at{=}0$ at the
write-in step: task-vector training still produces a per-task LoRA proposal, but the
proposal is not permanently accumulated into the running model.  We
use NoWrite only to attribute gains to task-vector shaping rather than
to write-in coefficient choices (Appendix~\ref{app:ablation}); it is
not a deployable continual method.

For an $\alpha$ grid $\mathcal{A}$ and seen-task accuracy curve
$\overline{\mathrm{Acc}}_{\le t}(\alpha)$, the path-level metrics are
\begin{align*}
R_t &= \max_{\alpha\in\mathcal{A}}\overline{\mathrm{Acc}}_{\le t}(\alpha)
       - \min_{\alpha\in\mathcal{A}}\overline{\mathrm{Acc}}_{\le t}(\alpha),
\\
W_{t,\varepsilon}
  &= \tfrac{1}{|\mathcal{A}|}\bigl|\{\alpha:
       \overline{\mathrm{Acc}}_{\le t}(\alpha)\geq
       \max_{\alpha'}\overline{\mathrm{Acc}}_{\le t}(\alpha')-\varepsilon\}\bigr|,
\\
G_{t,\mathrm{fix}}(\alpha_r)
  &= \max_{\alpha\in\mathcal{A}}\overline{\mathrm{Acc}}_{\le t}(\alpha)
       - \overline{\mathrm{Acc}}_{\le t}(\alpha_r).
\end{align*}
The tolerance $\varepsilon$ in $W_{t,\varepsilon}$ is in fractional
accuracy: $W_{t,0.0025}$ is the fraction of the grid within
$0.25$ percentage points of the optimum.  Cross-(seed, task)
aggregates always average over tasks within a seed first and then
report mean and standard deviation across seeds, so that task spread
does not leak into seed error bars.

\paragraph{A.6 Diagonal Fisher summary and task-vector statistics.}
For methods requiring Fisher information, we maintain a diagonal
Fisher summary over the adapted parameters following
\citet{qiu2025pm}.  The same stored summary is used for the
Fisher-derived coefficient in P\&M and \prm{}-Fisher and for the
diagnostic interference score $q_t = \langle \dt,\Fbar_{<t}\dt\rangle$
in Eq.~(\ref{eq:fisher-interference-main}); it is updated at task
boundaries and is not recomputed during the $\alpha$-sweep
diagnostics.  Each task-vector proposal additionally yields the
Frobenius norm $\|\dt\|_F$ and the direction-normalized interference
$q_t^{\mathrm{dir}}{=}q_t/(\|\dt\|_F^{2}+\varepsilon_{\mathrm{num}})$.
For non-negligible $\|\dt\|_F$, this gives the approximate decomposition
$q_t\approx\|\dt\|_F^{2}\,q_t^{\mathrm{dir}}$ used in
\S\ref{sec:method:metrics}.  The matched-prefix interventions in
Appendix~\ref{app:matched-prefix} freeze a single shared prefix and
train two twin task-vector proposals (P\&M and \prm{}) from identical
initialization, data order, and randomness, differing only in the
proximal weight; the resulting twin task vectors are then post-hoc
rescaled or norm-swapped and evaluated on the same $\alpha$ grid, so
the observed differences cannot be attributed to a different history
of previous merges.

\paragraph{A.7 Compute.}
All in-house experiments were run on NVIDIA A100-SXM4 GPUs with 80 GB memory,using one visible GPU per seed. A full PRM run takes approximately 2.5 GPU-hours for ImageNet-R 10T, 2.4 GPU-hours for CIFAR-100 10T, 2.6 GPU-hours for DomainNet 5T, and 0.4 GPU-hours for CUB-200 10T. The total compute for the reported
in-house experiments was approximately 625 GPU-hours. Preliminary sweeps and failed or aborted runs required approximately 26 additional GPU-hours, and therefore did not require substantially more compute than the reported experiments.

\paragraph{A.8 Baseline implementation details.}
Each baseline is described below with its origin, the principle of
its update rule, and whether the result was produced by our own
implementation under the protocol of paragraphs~A.1--A.2 or by
running the authors' released codebase.  Lifted numbers (marked
$\dagger$ in result tables) reuse the protocol of
\citet{qiu2025pm}.

\textbf{Full-FT.}
\emph{In-house.}  Standard joint fine-tuning of the entire backbone
on each task; no merging is performed and no parameter is shared
across tasks beyond the running model itself.  We used
Full-FT only as a non-merging reference; the dagger-marked Full-FT
cells are lifted from \citet{qiu2025pm}.

\textbf{LoRA.}
\emph{In-house.}  A per-task LoRA adapter is trained with the
optimizer settings of paragraph~A.2 and written into the running
model with $\at{=}1$.  No perturbation, no proximal term, no
Fisher.  This is the simplest naive sequential-LoRA baseline.

\textbf{EWC-LoRA.}
\emph{In-house, following \citet{kirkpatrick2017ewc} adapted to
LoRA.}  A \emph{single} shared LoRA adapter is reused across tasks
(the per-task allocation step in Algorithm~\ref{alg:prm} is skipped),
and at task $t$ the loss is
$\mathcal{L}_{\mathrm{CE}} + \tfrac{\lambda_{\mathrm{ewc}}}{2}\sum_{s<t}\sum_{\theta}F_s[\theta](\theta-\theta_s^{*})^{2}$,
where $\theta_s^{*}$ is the snapshot of the shared LoRA parameters
after task $s$ and $F_s$ is the per-task diagonal Fisher computed at
$\theta_s^{*}$.  No additional consolidation is applied; the model
at task $t$ is the EWC-regularised solution itself.

\textbf{Inf-LoRA \citep{liang2024inflora}.}
\emph{Lifted only.}  All Inf-LoRA cells in
Tables~\ref{tab:main-inr},~\ref{tab:main-other} are taken from
\citet{qiu2025pm}'s comparison tables, which in turn cite the numbers
reported in the original Inf-LoRA paper.  Those numbers were obtained
under the Inf-LoRA paper's own protocol, namely the
\texttt{vit\_base\_patch16\_224\_in21k} backbone (no AugReg ImageNet-1k
fine-tune), the upstream class ordering (alphabetical class folder
order, \texttt{shuffle}{=}\texttt{false}), and the upstream epoch
budget (50 epochs/task on ImageNet-R, 20 epochs/task on CIFAR-100).
We did not re-run Inf-LoRA in our protocol.  Aligning its DualGPM
subspace projection to a common AugReg backbone with $10$ epochs/task
without changing the algorithm is non-trivial: the SVD-based projector
initialization in \citet{liang2024inflora} depends on the specific
backbone's feature statistics, and the upstream learning-rate schedule
was tuned to the upstream epoch budget.  We therefore follow
\citet{qiu2025pm}'s convention of reporting Inf-LoRA's published
numbers as a reference rather than as a strictly protocol-matched
cell, and we mark every such cell with~$\dagger$.

\textbf{SD-LoRA \citep{wu2024sdlora}.}
\emph{Lifted only.}  All SD-LoRA cells in
Tables~\ref{tab:main-inr},~\ref{tab:main-other} are likewise taken
from \citet{qiu2025pm}'s tables, which in turn cite the original
SD-LoRA paper.  Those numbers use SD-LoRA's published protocol with
the \texttt{vit\_base\_patch16\_224} (ImageNet-1k pretrain) backbone
and the upstream optimizer / epoch budget, neither of which matches
our AugReg + 10-epoch setup.  We did not re-run SD-LoRA; cells are
marked with~$\dagger$.

\textbf{L2P, DualPrompt, CODA-Prompt.}
\emph{Lifted only}; we did not run these methods ourselves.  Cells
appear in the main results tables only when a $\dagger$-marked cell
is available from \citet{qiu2025pm} under the matching protocol.

\textbf{P\&M \citep{qiu2025pm}.}
\emph{In-house re-implementation.}  Per-task LoRA training with the
perturbation objective of \citet{qiu2025pm}: at each step the
forward pass is taken at
$\theta_{t-1}+(1{+}\tilde{\epsilon})B_tA_t$ with
$\tilde{\epsilon}\!\sim\!\mathcal{P}_{\epsilon}$ (zero with
probability $1{-}p_0$), and the cross-entropy loss is back-propagated
through the perturbed forward.  At the end of task-vector training, the
Fisher-derived coefficient $\at$ from Step~3 of
Algorithm~\ref{alg:prm} is used for write-in.  Our re-implementation
matches the published P\&M protocol to within seed noise on
ImageNet-R.

\textbf{P\&M-fixed.}
\emph{In-house.}  Same task-vector training as P\&M; the write-in rule replaces the Fisher
coefficient by a constant $\at{=}0.8$ across all tasks.

\textbf{Model-Avg.}
\emph{In-house re-implementation.}  Sequential model averaging:
treating the per-task LoRA proposal as $\theta_t^{*}{=}\theta_{t-1}+\dt$,
the running mean update
$\theta_t = \tfrac{t}{t+1}\theta_{t-1}+\tfrac{1}{t+1}\theta_t^{*}$
collapses to the scalar schedule $\at{=}1/(t{+}1)$ in our scalar
write-in interface.  We use the in-house P\&M task-vector training for $\dt$ and
only change the write-in schedule.

\textbf{CoMA \citep{marouf2024weighted}.}
\emph{In-house re-implementation.}  Continual model averaging: the
running mean form above with a constant scalar gate.  Task-vector training is the
same as P\&M; the write-in rule uses the CoMA scalar schedule.

\textbf{CoFiMA \citep{marouf2024weighted}.}
\emph{In-house re-implementation.}  Fisher-weighted continual model
averaging.  In our LoRA setting the per-element CoFiMA gate
\(g[j] = (a F_t[j])/((1-a)F_{<t}[j] + a F_t[j])\) (with $a$ a
hyper-parameter) replaces the scalar schedule of CoMA, applied
elementwise to $\dt$ before write-in.  Task-vector training is the same as P\&M.

\textbf{BECAME \citep{li2025became}.}
\emph{In-house re-implementation.}  Closed-form scalar coefficient
$\alpha^{*}{=}\sum_j F_t[j]\,\delta[j]^{2}\big/\sum_j (F_t[j]+F_{<t}[j])\,\delta[j]^{2}$,
where $F_t,F_{<t}$ are the diagonal Fishers on the K and V row
blocks, and $\delta=\dt$.  Task-vector training is the same as P\&M.  We followed
the closed-form merge derivation in the BECAME public release
(\url{https://github.com/limei0818/BECAME}).

\textbf{MagMax \citep{marczak2024magmax}.}
\emph{In-house re-implementation.}  Per-coordinate
maximum-magnitude task-vector merging: at each layer and each
coordinate $j$, the merged delta picks
$\dt[j]$ from the task with the largest $|\dt[j]|$.  We realise this
in the scalar write-in interface by setting $\at{=}1$ together with a
per-coordinate gate that is one-hot in the winning task and zero
elsewhere; the gate of every past task is recomputed at every
write-in because a coordinate may change owner.  Task-vector training is the same
as P\&M.

\textbf{\prm{} (this work).}
\emph{In-house.}  Algorithm~\ref{alg:prm} with
$\lambda_{\mathrm{prox}}{=}10^{-2}$.  \prm{}-fixed and \prm{}-Fisher
share task-vector training and differ only in Step~3.

\textbf{$+$Prox (rule).}
\emph{In-house.}  It uses P\&M-style task-vector training with the proximal penalty added;
the write-in step uses the merge rule of the named
baseline (CoFiMA, CoMA, BECAME, MagMax, Model-Avg, or
P\&M-fixed/P\&M).  Used in the universality study
(Appendix~\ref{app:universality}).

\paragraph{A.9 Datasets and class-incremental splits.}
We use four image datasets, all under the standard class-incremental
protocol: each class appears in exactly one task and the
class-to-task assignment is fixed before training.

\textbf{ImageNet-R \citep{hendrycks2021imagenetr}.}
$200$ classes drawn from artistic renditions, sketches, paintings,
and other domain shifts of ImageNet categories.  We use the
publicly released ImageNet-R split as redistributed by
\citet{qiu2025pm} for direct protocol compatibility, with the
$200$ classes re-grouped into $T\!\in\!\{5,10,20\}$ tasks of
$\{40,20,10\}$ classes each.

\textbf{CIFAR-100 \citep{krizhevsky2009cifar}.}
$100$ image classes at $32{\times}32$ resolution.  Images are
upsampled to the backbone's $224{\times}224$ input size with bilinear
interpolation.  We follow the class-incremental split protocol used
in \citet{qiu2025pm}, with $T\!\in\!\{10,20\}$ tasks containing
$\{10,5\}$ classes each.

\textbf{DomainNet (real domain) \citep{peng2019moment}.}
DomainNet is a domain-generalisation dataset with six domains; we
use only the \emph{real} domain restricted to its
$200$ most-populous classes, matching the ``S-DomainNet'' split
introduced in the P\&M codebase.  We download the official
\texttt{real} archive from the M3SDA release page and split it into
$T{=}5$ tasks of $40$ classes each.

\textbf{CUB-200 \citep{wah2011caltech}.}
The Caltech-UCSD Birds-200 dataset with $200$ fine-grained bird
species.  We use the standard train/test split of the public release,
re-grouped into $T{=}10$ tasks of $20$ classes each.

\textbf{Class permutation per seed.}
For each (dataset, $T$, seed) triple, the $200$ (or $100$) classes
are permuted by a PCG64 random number generator instantiated from
the seed.  The first $C/T$ classes are assigned to task~$1$, the
next $C/T$ to task~$2$, and so on, where $C$ is the number of
classes.  The same permutation is reused across every method
compared in-house, so that paired (seed, task) cells are directly
comparable.  Lifted-from-Qiu-et-al.\ baselines are run under the
matching protocol of \citet{qiu2025pm}, which uses the same family
of permutations on the same class lists.

\textbf{Train/validation handling.}
Within each task we use the dataset's standard
training set for task-vector training and the dataset's standard
validation/test set restricted to the classes seen so far for
evaluation.  No samples from previous tasks' training sets are
revisited (rehearsal-free).  Pre-processing follows the timm
defaults associated with the AugReg or CLIP-OpenAI ViT-B/16
checkpoint used by the run; only the input resize and normalisation
constants differ between the two backbones.

\FloatBarrier

\section{Extended benchmark results on AugReg}
\label{app:extended-results}
\label{app:per-dataset}

Table~\ref{tab:main-other} reports the per-dataset breakdown referred
to in \S\ref{sec:res:main}: CIFAR-100 with 10/20 tasks, DomainNet 5T,
and CUB-200 10T.  Conventions match Table~\ref{tab:main-inr}.

\begin{table}[H]
\centering
\caption{Main results on CIFAR-100, DomainNet, and CUB-200 (ViT-B/16 AugReg).}
\label{tab:main-other}
\setlength{\tabcolsep}{4pt}
\renewcommand{\arraystretch}{1.05}
\resizebox{\textwidth}{!}{%
\begin{tabular}{lcccccccc}
\toprule
 & \multicolumn{4}{c}{CIFAR-100} & \multicolumn{2}{c}{DomainNet} & \multicolumn{2}{c}{CUB-200} \\
\cmidrule(lr){2-5} \cmidrule(lr){6-7} \cmidrule(lr){8-9}
 & \multicolumn{2}{c}{T=10} & \multicolumn{2}{c}{T=20} & \multicolumn{2}{c}{T=5} & \multicolumn{2}{c}{T=10} \\
\cmidrule(lr){2-3} \cmidrule(lr){4-5} \cmidrule(lr){6-7} \cmidrule(lr){8-9}
Method & Acc\,$\uparrow$ & AAA\,$\uparrow$ & Acc\,$\uparrow$ & AAA\,$\uparrow$ & Acc\,$\uparrow$ & AAA\,$\uparrow$ & Acc\,$\uparrow$ & AAA\,$\uparrow$ \\
\midrule
Full-FT & 69.49\,$\pm$\,.50$^{\dagger}$ & 80.35\,$\pm$\,.87$^{\dagger}$ & 51.21\,$\pm$\,1.26$^{\dagger}$ & 61.83\,$\pm$\,1.20$^{\dagger}$ & 51.46\,$\pm$\,.47$^{\dagger}$ & 67.08\,$\pm$\,1.13$^{\dagger}$ & 51.43\,$\pm$\,1.41$^{\dagger}$ & 69.74\,$\pm$\,.93$^{\dagger}$ \\
L2P & 83.18\,$\pm$\,1.20$^{\dagger}$ & 87.69\,$\pm$\,1.05$^{\dagger}$ & \underline{85.66\,$\pm$\,.16}$^{\dagger}$ & \textbf{90.60\,$\pm$\,.23}$^{\dagger}$ & 70.26\,$\pm$\,.25$^{\dagger}$ & 75.83\,$\pm$\,.98$^{\dagger}$ & 65.18\,$\pm$\,2.49$^{\dagger}$ & 76.12\,$\pm$\,1.27$^{\dagger}$ \\
BECAME & 87.25\,$\pm$\,.49 & 91.22\,$\pm$\,1.76 & 84.82\,$\pm$\,.48 & 89.89\,$\pm$\,1.35 & 83.47\,$\pm$\,1.21 & 87.54\,$\pm$\,1.52 & 69.36\,$\pm$\,1.03 & 76.95\,$\pm$\,.81 \\
CoFiMA & 86.90\,$\pm$\,.19 & 90.99\,$\pm$\,1.68 & 84.91\,$\pm$\,.10 & \underline{90.23\,$\pm$\,1.22} & 83.71\,$\pm$\,.95 & 87.62\,$\pm$\,1.17 & 69.43\,$\pm$\,1.15 & 77.10\,$\pm$\,1.55 \\
CoMA & 85.92\,$\pm$\,.26 & 90.38\,$\pm$\,2.06 & 83.46\,$\pm$\,.30 & 89.52\,$\pm$\,1.21 & 83.59\,$\pm$\,.67 & 87.63\,$\pm$\,1.12 & 70.65\,$\pm$\,1.85 & 79.05\,$\pm$\,.50 \\
MagMax & 84.97\,$\pm$\,.72 & 87.59\,$\pm$\,2.27 & 84.74\,$\pm$\,.43 & 88.02\,$\pm$\,1.99 & 80.20\,$\pm$\,.76 & 82.34\,$\pm$\,1.24 & 70.95\,$\pm$\,2.14 & 78.46\,$\pm$\,.75 \\
Model-Avg & 87.16\,$\pm$\,1.02 & 90.91\,$\pm$\,2.43 & 84.74\,$\pm$\,.89 & 89.82\,$\pm$\,1.63 & 83.45\,$\pm$\,.93 & 87.51\,$\pm$\,1.18 & 70.58\,$\pm$\,1.54 & 78.81\,$\pm$\,.93 \\
EWC-LoRA & 85.95\,$\pm$\,.23 & 89.54\,$\pm$\,1.33 & 81.79\,$\pm$\,.83 & 85.11\,$\pm$\,2.39 & 83.88\,$\pm$\,.36 & 85.27\,$\pm$\,.83 & 46.76\,$\pm$\,3.19 & 63.95\,$\pm$\,1.14 \\
Inf-LoRA & 86.75\,$\pm$\,.35$^{\dagger}$ & 91.72\,$\pm$\,.15$^{\dagger}$ & 82.18\,$\pm$\,.88$^{\dagger}$ & 88.35\,$\pm$\,.77$^{\dagger}$ & 71.59\,$\pm$\,.23$^{\dagger}$ & 78.29\,$\pm$\,.50$^{\dagger}$ & \underline{76.68\,$\pm$\,.57}$^{\dagger}$ & \underline{81.39\,$\pm$\,.14}$^{\dagger}$ \\
SD-LoRA & \underline{88.01\,$\pm$\,.31}$^{\dagger}$ & \textbf{92.54\,$\pm$\,.18}$^{\dagger}$ & -- & -- & 72.82\,$\pm$\,.37$^{\dagger}$ & 78.44\,$\pm$\,.66$^{\dagger}$ & \textbf{77.48\,$\pm$\,.20}$^{\dagger}$ & \textbf{85.59\,$\pm$\,.44}$^{\dagger}$ \\
LoRA & 73.46\,$\pm$\,.41 & 81.61\,$\pm$\,2.31 & 63.99\,$\pm$\,1.30 & 78.55\,$\pm$\,1.69 & 72.46\,$\pm$\,.51 & 77.70\,$\pm$\,.74 & 64.43\,$\pm$\,2.13 & 73.82\,$\pm$\,.40 \\
P\&M & 87.46\,$\pm$\,.56 & 91.23\,$\pm$\,1.85 & 84.85\,$\pm$\,.58 & 89.97\,$\pm$\,1.39 & 83.87\,$\pm$\,.84 & 87.90\,$\pm$\,1.02 & 72.73\,$\pm$\,1.87 & 80.54\,$\pm$\,.42 \\
\midrule
\textbf{PRM-fixed} & 88.10\,$\pm$\,.52 & 92.03\,$\pm$\,1.03 & 85.43\,$\pm$\,.68 & 89.65\,$\pm$\,1.72 & \underline{84.74\,$\pm$\,.46} & \textbf{89.33\,$\pm$\,.59} & 73.65\,$\pm$\,1.89 & 80.81\,$\pm$\,.43 \\
\textbf{PRM-Fisher} & \textbf{88.25\,$\pm$\,.40} & \underline{92.08\,$\pm$\,.93} & \textbf{85.87\,$\pm$\,.54} & 89.87\,$\pm$\,1.64 & \textbf{84.76\,$\pm$\,.40} & \underline{89.26\,$\pm$\,.52} & 74.07\,$\pm$\,1.66 & 81.04\,$\pm$\,.74 \\
\bottomrule
\end{tabular}%
}
\\[3pt]
{\footnotesize Each cell is mean$\pm$std over 3 in-house seeds; $\dagger$=lifted from Qiu et al.\ 2025 NeurIPS Tabs.~2/4/5/9/11 (matching protocol); cells lacking ``$\pm$''=single available seed; ``--''=data unavailable. \textbf{Best} per column bold; \underline{second best} underlined.}
\end{table}

Across these AugReg settings, \prm{} remains competitive with the
strongest sequential-merging baselines and improves over P\&M on most
settings, with the clearest gains on DomainNet 5T and CUB-200 10T.
Among sequential-merging baselines, \prm{}-fixed or \prm{}-Fisher
achieves the strongest or near-strongest \textsc{AAA} in most cells.
The lifted Inf-LoRA and SD-LoRA numbers are included for reference
under the matching protocol; since they use a different parameter-
isolation style, we avoid treating them as part of the same merging-rule
comparison.

\section{Backbone transfer: AugReg and CLIP}
\label{app:backbone}
\label{app:backbones}

The main paper reports AugReg in Table~\ref{tab:main-inr} and uses CLIP 
for the transfer diagnostics in \S\ref{sec:res:universality}.  Here we
collect paired ImageNet-R 10T comparisons where both backbones have Base
and $+$Prox runs (Table \ref{tab:backbone-paired}). The five-backbone radar further includes supervised ImageNet-21K, DINO~\citep{caron2021emerging}, and MAE~\citep{he2022masked} ViT checkpoints.

\begin{table}[htbp]
\centering
\caption{Backbone-paired ImageNet-R 10T comparison.  Forgetting reduction is
Forg(Base) $-$ Forg(+Prox), so positive means improvement.}
\label{tab:backbone-paired}
\resizebox{0.95\textwidth}{!}{%
\setlength{\tabcolsep}{4pt}
\renewcommand{\arraystretch}{1.05}
\begin{tabular}{l c rr r rr r}
\toprule
Write-in rule & Backbone & \multicolumn{2}{c}{\textsc{AAA}} & $\Delta$\textsc{AAA} & \multicolumn{2}{c}{\textsc{Forg}} & $\Delta$\textsc{Forg} \\
\cmidrule(lr){3-4} \cmidrule(lr){6-7}
 &  & Base & +Prox & & Base & +Prox & (reduction) \\
\midrule
P\&M ($\alpha{=}0.8$) & AugReg & 82.30\,$\pm$\,0.97 & 86.39\,$\pm$\,0.53 & +4.09 & 16.06\,$\pm$\,0.94 & 5.62\,$\pm$\,0.29 & +10.44 \\
P\&M ($\alpha{=}0.8$) & CLIP & 60.77\,$\pm$\,0.02 & 84.13\,$\pm$\,0.90 & +23.36 & 52.41\,$\pm$\,1.19 & 12.77\,$\pm$\,0.69 & +39.64 \\
P\&M (Fisher $\alpha$) & AugReg & 85.55\,$\pm$\,0.90 & 85.96\,$\pm$\,0.48 & +0.41 & 3.50\,$\pm$\,0.41 & 4.28\,$\pm$\,0.28 & -0.78 \\
P\&M (Fisher $\alpha$) & CLIP & 77.53\,$\pm$\,0.73 & 85.55\,$\pm$\,0.60 & +8.02 & 7.72\,$\pm$\,1.09 & 5.93\,$\pm$\,0.81 & +1.79 \\
\bottomrule
\end{tabular}
}
\end{table}

\begin{figure}[htbp]
\centering
\includegraphics[width=0.4\textwidth]{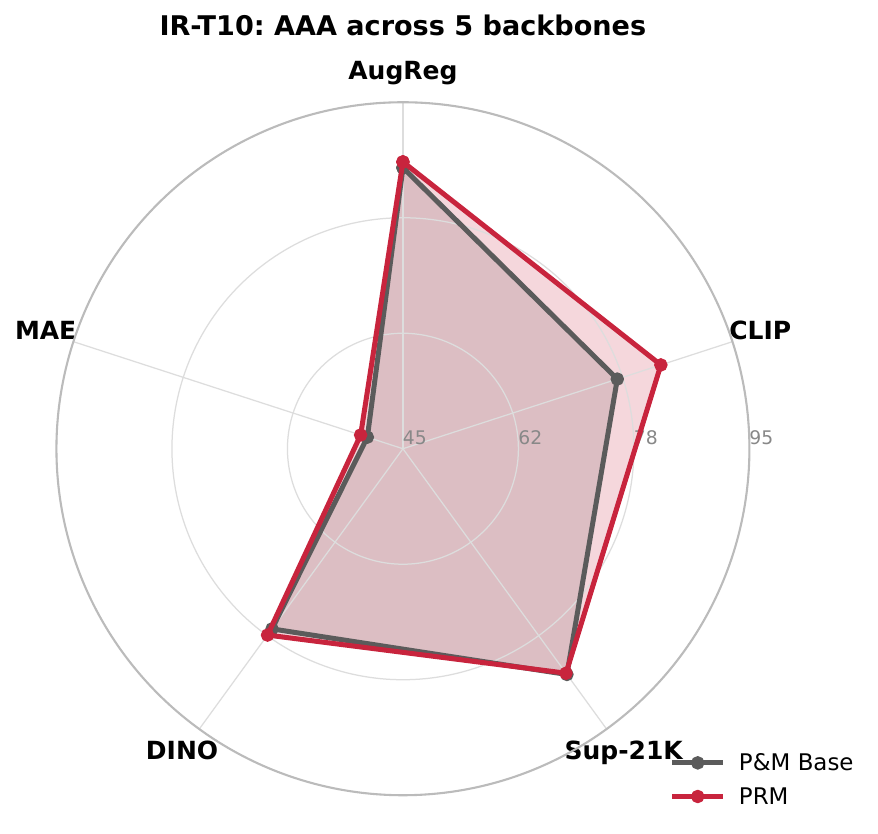}
\caption{Five-backbone radar on ImageNet-R 10T.  Each axis is one
backbone (AugReg, CLIP, Sup-21K, DINO, MAE) and the radial value is
\textsc{AAA}.  The two polylines compare P\&M Base against \prm{};
\prm{} improves over P\&M on every reported backbone, with the largest lift on
the backbones whose unshaped Base is the lowest.}
\label{fig:backbone-parallel}
\end{figure}

AugReg starts from a stronger Base than CLIP.  The gain from $+$Prox is
therefore smaller on AugReg but remains positive.  On the fixed
$\alpha{=}0.8$ rule, the \textsc{AAA} gain is $+4.09$ on AugReg and
$+23.36$ on CLIP; on the Fisher rule, the gain is $+0.41$ on AugReg and
$+8.02$ on CLIP.  The five-backbone radar in
Figure~\ref{fig:backbone-parallel} extends this view beyond AugReg
and CLIP to Sup-21K, DINO, and MAE, showing that \prm{} improves over P\&M
on every reported backbone, with the largest lifts on the backbones whose
unshaped Base \textsc{AAA} is the lowest.  This supports the view
that proximal shaping is most useful when the unregularized write-in
path is brittle.

\paragraph{Full CLIP main results.}
Tables~\ref{tab:clip-main-inr} and \ref{tab:clip-main-other} give the
full CLIP-backbone counterparts of Tables~\ref{tab:main-inr} and
\ref{tab:main-other}, restricted to the merge-baseline rows we ran
in-house on CLIP (BECAME, CoFiMA, CoMA, MagMax, Model-Avg, P\&M).  We
did not run Inf-LoRA, SD-LoRA, L2P, EWC-LoRA, LoRA, or Full-FT on
CLIP, so those rows are omitted; cells where a particular merge
baseline did not finish are rendered as ``--''.  Both \prm{}-fixed
and \prm{}-Fisher achieve the best \textsc{AAA} among the in-house CLIP merge baselines in every reported CLIP cell: on the
seven (dataset, $T$) settings, \prm{}-Fisher beats the strongest
in-house CLIP merge baseline by between $4.7$ and $11.3$ pp, and
\prm{}-fixed by between $3.4$ and $9.7$ pp.  The largest gains land
on the harder splits (CIFAR-100 T=20, IN-R T=20) where the unshaped
P\&M write-in is the most brittle, consistent with the
applicability pattern in Appendix~\ref{app:applicability}.

\begin{table*}[t]
\centering
\caption{Main results on ImageNet-R with the CLIP-OpenAI ViT-B/16 backbone. PRM-fixed and PRM-Fisher (bold rows) are our methods.}
\label{tab:clip-main-inr}
\setlength{\tabcolsep}{4pt}
\renewcommand{\arraystretch}{1.05}
\resizebox{\textwidth}{!}{%
\begin{tabular}{lcccccc}
\toprule
 & \multicolumn{6}{c}{ImageNet-R} \\
\cmidrule(lr){2-7}
 & \multicolumn{2}{c}{T=5} & \multicolumn{2}{c}{T=10} & \multicolumn{2}{c}{T=20} \\
\cmidrule(lr){2-3} \cmidrule(lr){4-5} \cmidrule(lr){6-7}
Method & Acc\,$\uparrow$ & AAA\,$\uparrow$ & Acc\,$\uparrow$ & AAA\,$\uparrow$ & Acc\,$\uparrow$ & AAA\,$\uparrow$ \\
\midrule
BECAME & 75.20\,$\pm$\,1.51 & 80.18\,$\pm$\,1.80 & 69.68\,$\pm$\,0.76 & 77.58\,$\pm$\,0.34 & 61.35\,$\pm$\,1.79 & 72.24\,$\pm$\,2.83 \\
CoFiMA & 72.92\,$\pm$\,0.85 & 78.92\,$\pm$\,1.62 & 65.00\,$\pm$\,1.04 & 75.53\,$\pm$\,0.39 & 51.82\,$\pm$\,1.35 & 68.76\,$\pm$\,1.14 \\
CoMA & 72.47\,$\pm$\,0.71 & 78.78\,$\pm$\,1.63 & 63.54\,$\pm$\,1.39 & 74.96\,$\pm$\,0.28 & 48.43\,$\pm$\,0.74 & 66.63\,$\pm$\,1.25 \\
MagMax & 62.01\,$\pm$\,0.52 & 67.81\,$\pm$\,1.05 & 58.86\,$\pm$\,0.66 & 65.66\,$\pm$\,0.26 & 55.29\,$\pm$\,0.93 & 64.67\,$\pm$\,1.24 \\
Model-Avg & 75.41\,$\pm$\,1.08 & 79.94\,$\pm$\,1.67 & 70.04\,$\pm$\,0.87 & 77.75\,$\pm$\,0.58 & 61.05\,$\pm$\,1.69 & 72.08\,$\pm$\,2.64 \\
P\&M & 75.35\,$\pm$\,1.29 & 80.20\,$\pm$\,1.76 & 69.95\,$\pm$\,0.77 & 77.53\,$\pm$\,0.73 & 61.61\,$\pm$\,1.88 & 72.45\,$\pm$\,2.96 \\
\midrule
\textbf{PRM-fixed} & \underline{81.86\,$\pm$\,1.10} & \underline{86.61\,$\pm$\,1.26} & \underline{77.19\,$\pm$\,0.43} & \underline{84.13\,$\pm$\,0.90} & \underline{66.49\,$\pm$\,0.05} & \underline{75.51\,$\pm$\,0.99} \\
\textbf{PRM-Fisher} & \textbf{82.41\,$\pm$\,0.74} & \textbf{87.36\,$\pm$\,0.63} & \textbf{78.71\,$\pm$\,0.75} & \textbf{85.55\,$\pm$\,0.60} & \textbf{70.67\,$\pm$\,1.85} & \textbf{77.88\,$\pm$\,1.27} \\
\bottomrule
\end{tabular}%
}
\\[3pt]
{\footnotesize Each cell is mean$\pm$std over 3 in-house seeds on the CLIP-OpenAI ViT-B/16 backbone. \textbf{Best} per column bold; \underline{second best} underlined.  Bold/underline are applied independently to \textsc{Acc} and \textsc{AAA}.}
\end{table*}

\begin{table*}[t]
\centering
\caption{Main results on CIFAR-100, DomainNet, and CUB-200 with the CLIP-OpenAI ViT-B/16 backbone.}
\label{tab:clip-main-other}
\setlength{\tabcolsep}{4pt}
\renewcommand{\arraystretch}{1.05}
\resizebox{\textwidth}{!}{%
\begin{tabular}{lcccccccc}
\toprule
 & \multicolumn{4}{c}{CIFAR-100} & \multicolumn{2}{c}{DomainNet} & \multicolumn{2}{c}{CUB-200} \\
\cmidrule(lr){2-5} \cmidrule(lr){6-7} \cmidrule(lr){8-9}
 & \multicolumn{2}{c}{T=10} & \multicolumn{2}{c}{T=20} & \multicolumn{2}{c}{T=5} & \multicolumn{2}{c}{T=10} \\
\cmidrule(lr){2-3} \cmidrule(lr){4-5} \cmidrule(lr){6-7} \cmidrule(lr){8-9}
Method & Acc\,$\uparrow$ & AAA\,$\uparrow$ & Acc\,$\uparrow$ & AAA\,$\uparrow$ & Acc\,$\uparrow$ & AAA\,$\uparrow$ & Acc\,$\uparrow$ & AAA\,$\uparrow$ \\
\midrule
BECAME & 69.86\,$\pm$\,1.41 & 78.57\,$\pm$\,1.60 & 58.56\,$\pm$\,3.17 & 68.13\,$\pm$\,3.88 & 80.89\,$\pm$\,1.10 & 83.74\,$\pm$\,1.22 & 43.97\,$\pm$\,1.69 & 54.36\,$\pm$\,3.01 \\
CoFiMA & 61.99\,$\pm$\,1.41 & 73.91\,$\pm$\,1.58 & 46.26\,$\pm$\,1.53 & 62.65\,$\pm$\,1.48 & 78.98\,$\pm$\,0.26 & 82.99\,$\pm$\,0.81 & 44.10\,$\pm$\,1.00 & 55.51\,$\pm$\,2.61 \\
CoMA & 58.15\,$\pm$\,2.39 & 72.12\,$\pm$\,1.77 & 36.08\,$\pm$\,2.57 & 56.31\,$\pm$\,1.16 & 78.16\,$\pm$\,0.23 & 82.74\,$\pm$\,0.81 & 39.49\,$\pm$\,0.24 & 53.80\,$\pm$\,2.51 \\
MagMax & 52.06\,$\pm$\,1.73 & 60.97\,$\pm$\,1.05 & 43.11\,$\pm$\,1.56 & 52.37\,$\pm$\,1.06 & 68.49\,$\pm$\,1.53 & 72.54\,$\pm$\,0.60 & 40.92\,$\pm$\,0.45 & 52.73\,$\pm$\,1.57 \\
Model-Avg & 68.15\,$\pm$\,0.46 & 76.12\,$\pm$\,2.02 & 57.35\,$\pm$\,4.67 & 67.22\,$\pm$\,4.37 & 81.26\,$\pm$\,0.55 & 83.90\,$\pm$\,1.13 & 45.79\,$\pm$\,1.49 & 56.04\,$\pm$\,2.52 \\
P\&M & 70.04\,$\pm$\,1.26 & 78.72\,$\pm$\,1.56 & 58.33\,$\pm$\,3.16 & 67.85\,$\pm$\,3.93 & 80.91\,$\pm$\,1.21 & 83.72\,$\pm$\,1.26 & 44.23\,$\pm$\,1.59 & 54.38\,$\pm$\,2.93 \\
\midrule
\textbf{PRM-fixed} & \underline{77.54\,$\pm$\,0.53} & \underline{84.77\,$\pm$\,1.00} & \underline{69.22\,$\pm$\,1.17} & \underline{76.51\,$\pm$\,0.62} & \underline{85.29\,$\pm$\,0.77} & \underline{88.69\,$\pm$\,1.21} & \underline{47.68\,$\pm$\,1.75} & \underline{58.05\,$\pm$\,2.14} \\
\textbf{PRM-Fisher} & \textbf{78.66\,$\pm$\,0.74} & \textbf{85.59\,$\pm$\,1.43} & \textbf{71.60\,$\pm$\,0.39} & \textbf{79.13\,$\pm$\,0.68} & \textbf{86.16\,$\pm$\,0.08} & \textbf{90.34\,$\pm$\,0.13} & \textbf{53.39\,$\pm$\,0.74} & \textbf{61.31\,$\pm$\,2.52} \\
\bottomrule
\end{tabular}%
}
\\[3pt]
{\footnotesize Each cell is mean$\pm$std over 3 in-house seeds on the CLIP-OpenAI ViT-B/16 backbone. \textbf{Best} per column bold; \underline{second best} underlined.  Bold/underline are applied independently to \textsc{Acc} and \textsc{AAA}.}
\end{table*}

\section{Transfer beyond P\&M}
\label{app:universality}
\label{app:beyond-pm}

Appendix~\ref{app:universality} tests whether the same proximal task-vector-training
modification transfers beyond the P\&M write-in coefficient rule.  For each
merge rule, Base denotes the original training recipe and $+$Prox adds
$\lambda_{\mathrm{prox}}{=}10^{-2}$ while leaving the write-in rule
unchanged.

The full per-cell tables are reported in Tables~\ref{tab:beyond-pm-inr10}--\ref{tab:beyond-pm-cub200-T10}.  Conventions match
Table~\ref{tab:beyond-pm-inr10}.  Cells marked with $\dagger$ aggregate
a single available seed and should be interpreted as directional
checks rather than fully powered estimates.

\begin{table}[H]
\centering
\caption{Beyond P\&M generality on CLIP IN-R 10T (3 seeds, mean$\pm$std). Each write-in rule paired with no-prox \texttt{Base} and +Prox ($\lambda_{\mathrm{prox}}=10^{-2}$). $\Delta\textsc{AAA}$ and $\Delta\textsc{Forg}$ are computed as +Prox minus Base; positive $\Delta\textsc{AAA}$ and negative $\Delta\textsc{Forg}$ both indicate improvement. In this setting, every reported rule receives positive $\Delta\textsc{AAA}$ and negative $\Delta\textsc{Forg}$. Bold marks improvements.}
\label{tab:beyond-pm-inr10}
\setlength{\tabcolsep}{3pt}
\renewcommand{\arraystretch}{1.0}
\small
\begin{tabular}{l rrr rrr}
\toprule
Merge rule & \multicolumn{3}{c}{AAA $\uparrow$} & \multicolumn{3}{c}{Forgetting $\downarrow$} \\
\cmidrule(lr){2-4}\cmidrule(lr){5-7}
 & Base & +Prox & $\Delta$ & Base & +Prox & $\Delta$ \\
\midrule
P\&M (Fisher $\alpha$) & 77.53\,$\pm$\,0.73 & 85.55\,$\pm$\,0.60 & \textbf{+8.02} & 7.72\,$\pm$\,1.09 & 5.93\,$\pm$\,0.81 & \textbf{-1.79} \\
P\&M ($\alpha{=}0.8$ fixed) & 60.77\,$\pm$\,0.02 & 84.13\,$\pm$\,0.90 & \textbf{+23.36} & 52.41\,$\pm$\,1.19 & 12.77\,$\pm$\,0.69 & \textbf{-39.64} \\
CoFiMA & 75.53\,$\pm$\,0.39 & 85.77\,$\pm$\,0.55 & \textbf{+10.24} & 26.13\,$\pm$\,1.64 & 9.14\,$\pm$\,0.77 & \textbf{-16.99} \\
CoMA & 74.96\,$\pm$\,0.28 & 85.75\,$\pm$\,0.55 & \textbf{+10.79} & 28.70\,$\pm$\,2.15 & 9.23\,$\pm$\,0.95 & \textbf{-19.47} \\
Model-Avg & 77.75\,$\pm$\,0.58 & 85.74\,$\pm$\,0.58 & \textbf{+7.99} & 8.75\,$\pm$\,0.64 & 5.69\,$\pm$\,0.77 & \textbf{-3.06} \\
BECAME & 77.58\,$\pm$\,0.34 & 85.70\,$\pm$\,0.56 & \textbf{+8.12} & 8.35\,$\pm$\,0.96 & 5.75\,$\pm$\,0.71 & \textbf{-2.60} \\
MagMax & 65.66\,$\pm$\,0.31 & 83.42\,$\pm$\,1.28 & \textbf{+17.76} & 34.68\,$\pm$\,1.72 & 9.55\,$\pm$\,0.57 & \textbf{-25.12} \\

\bottomrule
\end{tabular}
\end{table}
\begin{table}[H]
\centering
\caption{Beyond P\&M generality on CLIP IN-R 20T (3 seeds, mean$\pm$std). Same protocol as Table~\ref{tab:beyond-pm-inr10}; harder split with twice the merge events. $\Delta$ columns are +Prox minus Base; positive $\Delta\textsc{AAA}$ and negative $\Delta\textsc{Forg}$ indicate improvement. Positive $\Delta\textsc{AAA}$ persists across all 7 rules; forgetting changes are rule-dependent, with the largest reductions on rules whose Base over-writes (e.g.~$\alpha{=}0.8$ fixed).}
\label{tab:beyond-pm-inr20}
\setlength{\tabcolsep}{3pt}
\renewcommand{\arraystretch}{1.0}
\small
\begin{tabular}{l rrr rrr}
\toprule
Merge rule & \multicolumn{3}{c}{AAA $\uparrow$} & \multicolumn{3}{c}{Forgetting $\downarrow$} \\
\cmidrule(lr){2-4}\cmidrule(lr){5-7}
 & Base & +Prox & $\Delta$ & Base & +Prox & $\Delta$ \\
\midrule
P\&M (Fisher $\alpha$) & 72.45\,$\pm$\,2.96 & 77.88\,$\pm$\,1.27 & \textbf{+5.44} & 6.26\,$\pm$\,0.68 & 7.42\,$\pm$\,1.12 & +1.16 \\
P\&M ($\alpha{=}0.8$ fixed) & 49.49\,$\pm$\,0.80 & 75.51\,$\pm$\,0.99 & \textbf{+26.02} & 67.48\,$\pm$\,1.52 & 20.71\,$\pm$\,1.36 & \textbf{-46.77} \\
CoFiMA & 68.76\,$\pm$\,1.14 & 78.14\,$\pm$\,1.24 & \textbf{+9.38} & 33.41\,$\pm$\,1.19 & 14.70\,$\pm$\,0.96 & \textbf{-18.71} \\
CoMA & 66.63\,$\pm$\,1.25 & 78.04\,$\pm$\,1.31 & \textbf{+11.41} & 40.16\,$\pm$\,1.21 & 14.52\,$\pm$\,1.03 & \textbf{-25.64} \\
Model-Avg & 72.08\,$\pm$\,2.64 & 77.82\,$\pm$\,1.52 & \textbf{+5.74} & 6.59\,$\pm$\,0.84 & 7.45\,$\pm$\,1.12 & +0.85 \\
BECAME & 72.24\,$\pm$\,2.83 & 77.90\,$\pm$\,1.28 & \textbf{+5.66} & 6.47\,$\pm$\,0.99 & 7.41\,$\pm$\,1.06 & +0.94 \\
MagMax & 64.67\,$\pm$\,1.52 & 77.26\,$\pm$\,1.12 & \textbf{+12.59} & 32.69\,$\pm$\,1.81 & 12.28\,$\pm$\,1.49 & \textbf{-20.41} \\

\bottomrule
\end{tabular}
\end{table}
\begin{table}[H]
\centering
\caption{Beyond P\&M generality on CLIP CIFAR-100 20T (CLIP backbone). $\dagger$: cell aggregates a single available seed; standard deviation is omitted.}
\label{tab:beyond-pm-cifar100-T20}
\setlength{\tabcolsep}{3pt}
\renewcommand{\arraystretch}{1.0}
\small
\begin{tabular}{l rrr rrr}
\toprule
Merge rule & \multicolumn{3}{c}{AAA $\uparrow$} & \multicolumn{3}{c}{Forgetting $\downarrow$} \\
\cmidrule(lr){2-4}\cmidrule(lr){5-7}
 & Base & +Prox & $\Delta$ & Base & +Prox & $\Delta$ \\
\midrule
P\&M (Fisher $\alpha$) & 67.85\,$\pm$\,3.93 & 79.13\,$\pm$\,0.68 & \textbf{+11.28} & 8.66\,$\pm$\,2.72 & 9.33\,$\pm$\,0.74 & +0.67 \\
P\&M ($\alpha{=}0.8$ fixed) & 34.81\,$\pm$\,2.05 & 76.51\,$\pm$\,0.62 & \textbf{+41.70} & 82.84\,$\pm$\,1.24 & 19.80\,$\pm$\,1.17 & \textbf{-63.04} \\
CoFiMA & 62.65\,$\pm$\,1.48 & 78.94\,$\pm$\,0.59 & \textbf{+16.29} & 39.94\,$\pm$\,3.20 & 15.13\,$\pm$\,0.66 & \textbf{-24.81} \\
CoMA & 56.31\,$\pm$\,1.16 & 79.53$^{\dagger}$ & \textbf{+23.22} & 58.26\,$\pm$\,5.10 & 15.60$^{\dagger}$ & \textbf{-42.66} \\
Model-Avg & 67.22\,$\pm$\,4.37 & 79.13\,$\pm$\,0.93 & \textbf{+11.91} & 8.91\,$\pm$\,2.75 & 9.26\,$\pm$\,0.57 & +0.35 \\
BECAME & 68.13\,$\pm$\,3.88 & 79.12\,$\pm$\,0.74 & \textbf{+11.00} & 8.72\,$\pm$\,2.82 & 9.40\,$\pm$\,0.64 & +0.67 \\
MagMax & 52.37\,$\pm$\,1.06 & 77.92\,$\pm$\,1.19 & \textbf{+25.54} & 50.49\,$\pm$\,2.68 & 14.43\,$\pm$\,0.52 & \textbf{-36.07} \\
\bottomrule
\end{tabular}
\end{table}

\begin{table}[H]
\centering
\caption{Beyond P\&M generality on CLIP DomainNet 5T (CLIP backbone).}
\label{tab:beyond-pm-domainnet-T5}
\setlength{\tabcolsep}{3pt}
\renewcommand{\arraystretch}{1.0}
\small
\begin{tabular}{l rrr rrr}
\toprule
Merge rule & \multicolumn{3}{c}{AAA $\uparrow$} & \multicolumn{3}{c}{Forgetting $\downarrow$} \\
\cmidrule(lr){2-4}\cmidrule(lr){5-7}
 & Base & +Prox & $\Delta$ & Base & +Prox & $\Delta$ \\
\midrule
P\&M (Fisher $\alpha$) & 83.72\,$\pm$\,1.26 & 90.34\,$\pm$\,0.13 & \textbf{+6.62} & 11.44\,$\pm$\,0.82 & 6.87\,$\pm$\,0.53 & \textbf{-4.58} \\
P\&M ($\alpha{=}0.8$ fixed) & 72.30\,$\pm$\,0.21 & 88.69\,$\pm$\,1.21 & \textbf{+16.39} & 37.74\,$\pm$\,0.05 & 9.27\,$\pm$\,1.11 & \textbf{-28.46} \\
CoFiMA & 82.99\,$\pm$\,0.81 & 90.34\,$\pm$\,0.39 & \textbf{+7.35} & 18.48\,$\pm$\,0.54 & 7.55\,$\pm$\,0.67 & \textbf{-10.93} \\
CoMA & 82.74\,$\pm$\,0.81 & 90.09\,$\pm$\,0.21 & \textbf{+7.35} & 19.97\,$\pm$\,0.53 & 7.75\,$\pm$\,0.73 & \textbf{-12.22} \\
Model-Avg & 83.90\,$\pm$\,1.13 & 90.34\,$\pm$\,0.38 & \textbf{+6.45} & 12.02\,$\pm$\,0.54 & 7.08\,$\pm$\,0.66 & \textbf{-4.94} \\
BECAME & 83.74\,$\pm$\,1.22 & 90.26\,$\pm$\,0.08 & \textbf{+6.52} & 11.39\,$\pm$\,0.80 & 6.94\,$\pm$\,0.65 & \textbf{-4.45} \\
MagMax & 72.54\,$\pm$\,0.60 & 88.42\,$\pm$\,1.10 & \textbf{+15.88} & 33.20\,$\pm$\,1.88 & 9.15\,$\pm$\,0.94 & \textbf{-24.05} \\
\bottomrule
\end{tabular}
\end{table}

\begin{table}[H]
\centering
\caption{Beyond P\&M generality on CLIP CUB-200 10T (CLIP backbone).}
\label{tab:beyond-pm-cub200-T10}
\setlength{\tabcolsep}{3pt}
\renewcommand{\arraystretch}{1.0}
\small
\begin{tabular}{l rrr rrr}
\toprule
Merge rule & \multicolumn{3}{c}{AAA $\uparrow$} & \multicolumn{3}{c}{Forgetting $\downarrow$} \\
\cmidrule(lr){2-4}\cmidrule(lr){5-7}
 & Base & +Prox & $\Delta$ & Base & +Prox & $\Delta$ \\
\midrule
P\&M (Fisher $\alpha$) & 54.38\,$\pm$\,2.93 & 61.31\,$\pm$\,2.52 & \textbf{+6.94} & 14.70\,$\pm$\,1.50 & 14.40\,$\pm$\,2.26 & \textbf{-0.30} \\
P\&M ($\alpha{=}0.8$ fixed) & 46.11\,$\pm$\,1.69 & 58.05\,$\pm$\,2.14 & \textbf{+11.94} & 66.75\,$\pm$\,1.13 & 35.84\,$\pm$\,3.39 & \textbf{-30.91} \\
CoFiMA & 55.51\,$\pm$\,2.61 & 61.36\,$\pm$\,3.10 & \textbf{+5.85} & 27.14\,$\pm$\,1.77 & 25.27\,$\pm$\,2.64 & \textbf{-1.87} \\
CoMA & 53.80\,$\pm$\,2.51 & 61.64\,$\pm$\,2.61 & \textbf{+7.84} & 44.65\,$\pm$\,0.42 & 25.58\,$\pm$\,2.20 & \textbf{-19.07} \\
Model-Avg & 56.04\,$\pm$\,2.52 & 61.82\,$\pm$\,2.39 & \textbf{+5.78} & 16.95\,$\pm$\,1.53 & 15.15\,$\pm$\,2.42 & \textbf{-1.80} \\
BECAME & 54.36\,$\pm$\,3.01 & 61.32\,$\pm$\,2.48 & \textbf{+6.96} & 14.79\,$\pm$\,1.17 & 14.45\,$\pm$\,2.40 & \textbf{-0.34} \\
MagMax & 52.73\,$\pm$\,1.57 & 59.48\,$\pm$\,1.85 & \textbf{+6.75} & 44.26\,$\pm$\,0.68 & 23.39\,$\pm$\,0.77 & \textbf{-20.87} \\
\bottomrule
\end{tabular}
\end{table}

\FloatBarrier
\paragraph{Cross-setting summary.}
Table~\ref{tab:universality-summary} aggregates the per-cell results by
setting.  Across the available CLIP settings, $+$Prox improves
\textsc{AAA} for every reported merge rule.  The largest mean gains
occur in the settings where the Base models are most sensitive to the
write-in coefficient.

\begin{table}[htbp]
\centering
\caption{Per-setting transfer summary on CLIP.  ``Rules
improved/total'' counts the available merge rules with positive
$\Delta\textsc{AAA}$ from Base to $+$Prox.  Forgetting reduction is
Forg(Base) $-$ Forg(+Prox), so positive means improvement.}
\label{tab:universality-summary}
\resizebox{\textwidth}{!}{%
\setlength{\tabcolsep}{4pt}
\renewcommand{\arraystretch}{1.05}
\begin{tabular}{l c r r r r}
\toprule
Setting & Rules \emph{improved}/\emph{total} (AAA) & Mean $\Delta$\textsc{AAA} & Median $\Delta$\textsc{AAA} & Best $\Delta$\textsc{AAA} & Mean Forg.\ reduction \\
\midrule
CLIP IN-R 10T & 7/7 & +12.33 & +10.24 & +23.36 & +15.52 \\
CLIP IN-R 20T & 7/7 & +10.89 & +9.38 & +26.02 & +15.51 \\
CLIP CIFAR-100 20T & 7/7 & +20.13 & +16.29 & +41.70 & +23.55 \\
CLIP DomainNet 5T & 7/7 & +9.51 & +7.35 & +16.39 & +12.80 \\
CLIP CUB-200 10T & 7/7 & +7.44 & +6.94 & +11.94 & +10.74 \\
\bottomrule
\end{tabular}
}
\end{table}

\FloatBarrier

\section{Matched-prefix scale interventions}
\label{app:matched-prefix}
\label{app:causal}  

Appendix~\ref{app:matched-prefix} isolates task-vector radius from
direction using matched-prefix interventions.  Both interventions start
from the same prefix model; the only difference is the task
vector trained from that prefix.

\subsection{Common-prefix protocol}
\label{app:matched-prefix:protocol}

We first run the unregularized P\&M pipeline on CLIP ImageNet-R 10T up
to the prefix before task~5.  From this common prefix, we train two
task-vector proposals with identical initialization, data order, and
randomness: one P\&M proposal with $\lambda_{\mathrm{prox}}{=}0$ and one
\prm{} proposal with $\lambda_{\mathrm{prox}}{=}10^{-2}$.  The two
proposals are then reloaded onto the same prefix for post-hoc
coefficient sweeps.  At every $\alpha\in\{0,0.05,\dots,1.0\}$, we
evaluate seen-task CE loss and accuracy on the held-out validation
splits and compute $R_t$, $W_{t,\varepsilon}$, and
$G_{t,\mathrm{fix}}$ as defined in Appendix~\ref{app:details}.

\subsection{Norm-matched task-vector intervention}
\label{app:matched-prefix:norm}

\begin{table}[htbp]
\centering
\caption{Norm-matched task-vector intervention on CLIP IN-R 10T (task~5).
Each row fixes the direction source and norm source before evaluating the
same common-prefix $\alpha$ sweep.}
\label{tab:norm-matched}
\resizebox{\textwidth}{!}{%
\setlength{\tabcolsep}{4pt}
\renewcommand{\arraystretch}{1.05}
\begin{tabular}{lcc rrrrrr}
\toprule
Vector & Direction & Norm & $\|\delta_t\|_F$ & $q_t$ [$10^{-3}$] & $q_t^{\mathrm{dir}}$ [$10^{-3}$] & $R_t$ [pp] & $G_{t,\mathrm{fix}}$ [pp] & \textsc{Acc}$_{\le t}{@}\alpha{=}0.8$ \\
\midrule
P\&M raw & P\&M & P\&M & 11.72\,$\pm$\,0.50 & 162.82\,$\pm$\,12.22 & 1.18\,$\pm$\,0.01 & 17.77\,$\pm$\,0.53 & 10.73\,$\pm$\,0.46 & 66.72\,$\pm$\,0.61 \\
P\&M $\!\to\!$ PRM-norm & P\&M & PRM & 1.56\,$\pm$\,0.02 & 2.88\,$\pm$\,0.11 & 1.18\,$\pm$\,0.01 & 2.37\,$\pm$\,0.44 & 0.33\,$\pm$\,0.11 & 76.49\,$\pm$\,0.09 \\
PRM raw & PRM & PRM & 1.56\,$\pm$\,0.02 & 29.91\,$\pm$\,0.51 & 12.29\,$\pm$\,0.30 & 3.07\,$\pm$\,0.82 & 0.27\,$\pm$\,0.19 & 77.25\,$\pm$\,0.55 \\
PRM $\!\to\!$ P\&M-norm & PRM & P\&M & 11.72\,$\pm$\,0.50 & 1694.62\,$\pm$\,177.89 & 12.29\,$\pm$\,0.30 & 70.01\,$\pm$\,0.49 & 63.28\,$\pm$\,0.74 & 14.05\,$\pm$\,1.31 \\
\bottomrule
\end{tabular}
}
\end{table}

\begin{figure}[htbp]
\centering

\begin{subfigure}{0.6\textwidth}
\centering
\includegraphics[width=\linewidth,height=0.2\textheight,keepaspectratio]{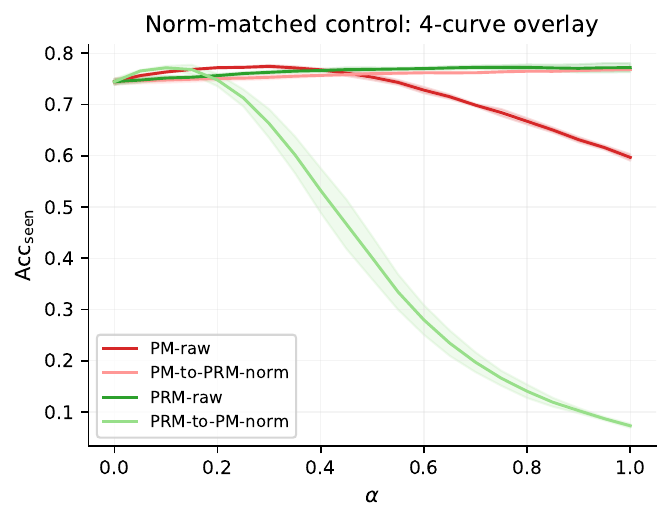}
\caption{$\overline{\mathrm{Acc}}_{\le t}(\alpha)$, four-vector overlay.}
\label{fig:norm-matched-a}
\end{subfigure}

\vspace{0.6em}

\begin{subfigure}{0.95\textwidth}
\centering
\includegraphics[width=\linewidth,height=0.28\textheight,keepaspectratio]{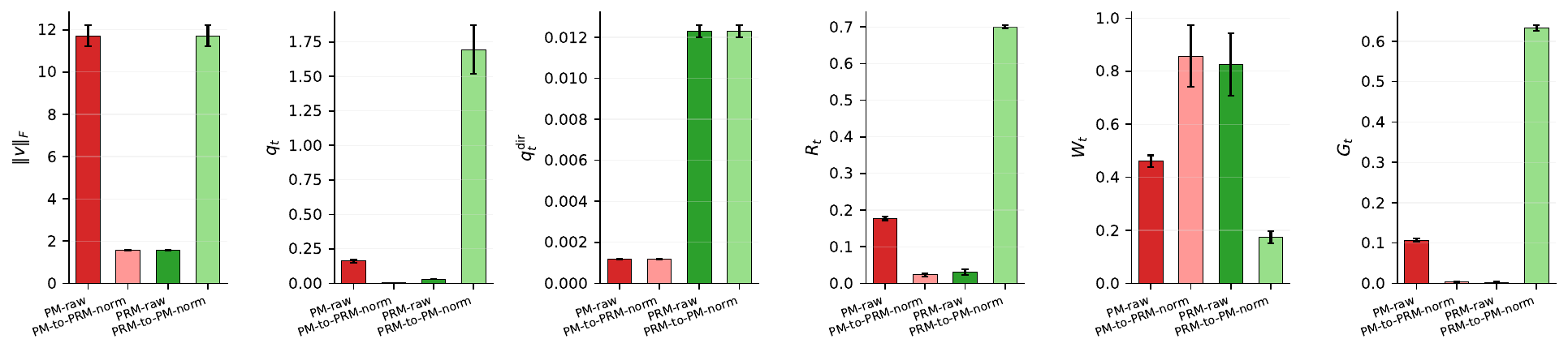}
\caption{Path metrics ($R_t$, $q_t$, $G_{t,\mathrm{fix}}$).}
\label{fig:norm-matched-b}
\end{subfigure}

\caption{Norm-matched intervention on CLIP IN-R 10T (task~5).
Rescaling $\delta_{\textsc{P\&M}}$ to \prm{}'s norm without altering
its direction substantially reduces coefficient sensitivity; inflating
$\delta_{\prm{}}$ to P\&M's norm substantially increases coefficient sensitivity.}
\label{fig:norm-matched}
\end{figure}

The four-vector table reports the source of direction and the source
of norm for each vector, and Figure~\ref{fig:norm-matched} visualizes
the corresponding $\alpha$-paths and path metrics.  Rescaling the P\&M
vector to the \prm{} norm preserves its direction but reduces its radius
by approximately $7.5\times$; its $R_t$ drops from $17.77$ pp to
$2.37$ pp, and $G_{t,\mathrm{fix}}$ drops from $10.73$ pp to $0.33$ pp.
Conversely, inflating the \prm{} direction to the P\&M norm reduces
$\overline{\mathrm{Acc}}_{\le t}@\alpha{=}0.8$ from $77.25$ to $14.05$.
The direction-normalized interference is actually larger under \prm{},
so the lower total interference is explained mainly by the smaller
radius entering $q_t\approx\|\dt\|_F^2 q_t^{\mathrm{dir}}$ for non-negligible $\|\dt\|_F$.

\subsection{Continuous task-vector rescaling intervention}
\label{app:matched-prefix:scaling}

\begin{figure}[htbp]
\centering
\begin{subfigure}[t]{0.32\textwidth}
\centering
\includegraphics[width=\textwidth]{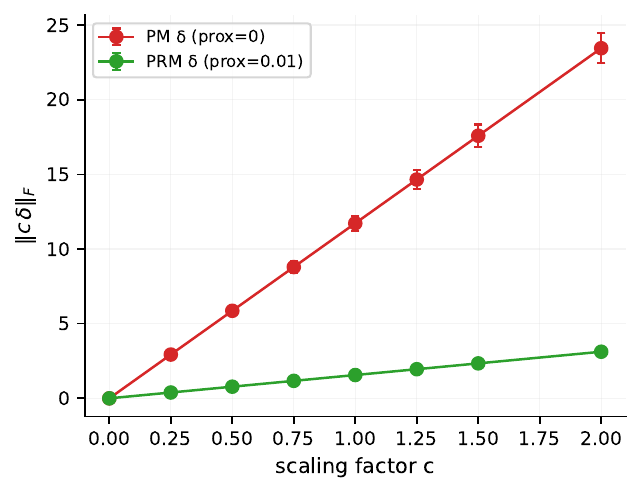}
\caption{$\|\delta(c)\|_F$ vs.\ $c$.}
\end{subfigure}
\hfill
\begin{subfigure}[t]{0.32\textwidth}
\centering
\includegraphics[width=\textwidth]{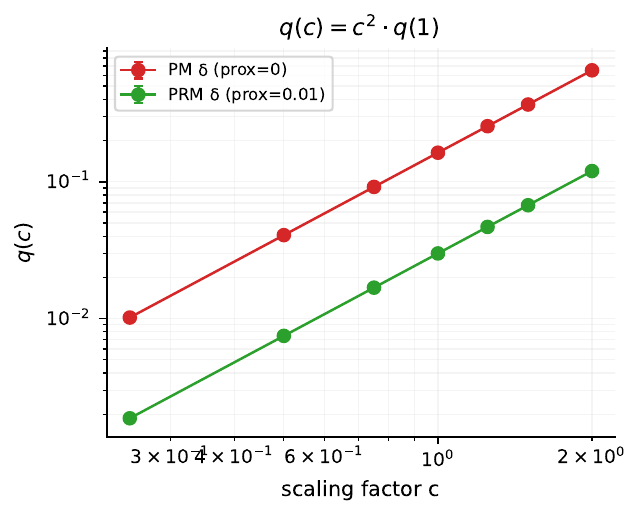}
\caption{$q_t(c)$ vs.\ $\|\delta(c)\|_F$.}
\end{subfigure}
\hfill
\begin{subfigure}[t]{0.32\textwidth}
\centering
\includegraphics[width=\textwidth]{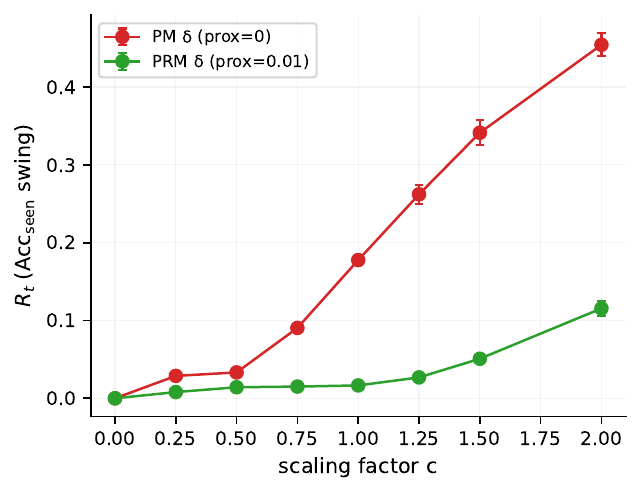}
\caption{$R_t(c)$ vs.\ $c$.}
\end{subfigure}

\begin{subfigure}[t]{0.32\textwidth}
\centering
\includegraphics[width=\textwidth]{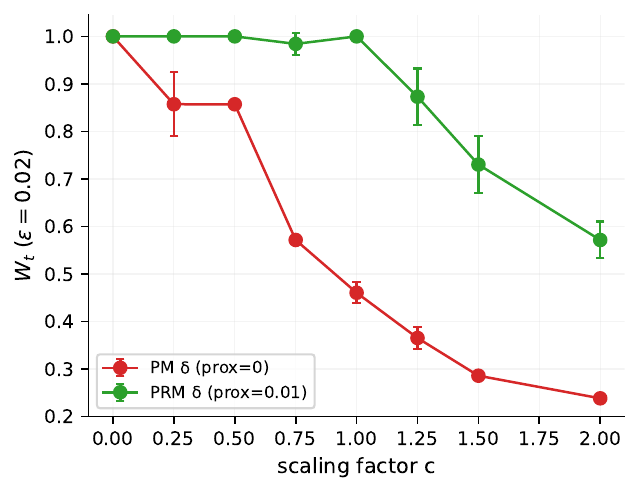}
\caption{$W_{t,0.005}(c)$ vs.\ $c$.}
\end{subfigure}
\hfill
\begin{subfigure}[t]{0.32\textwidth}
\centering
\includegraphics[width=\textwidth]{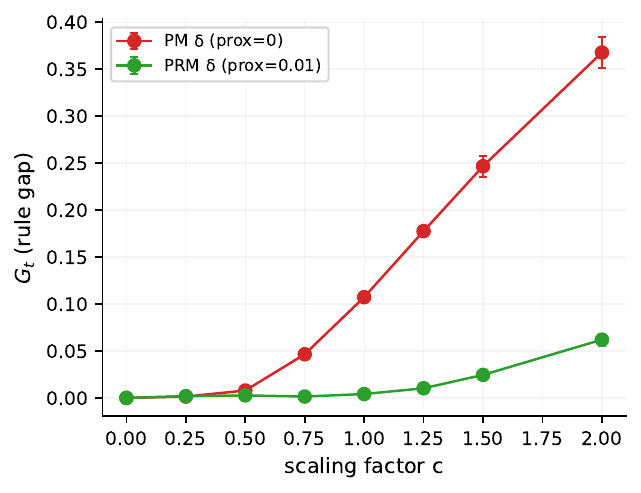}
\caption{$G_{t,\mathrm{fix}}(c)$ vs.\ $c$.}
\end{subfigure}
\hfill
\begin{subfigure}[t]{0.32\textwidth}
\centering
\includegraphics[width=\textwidth]{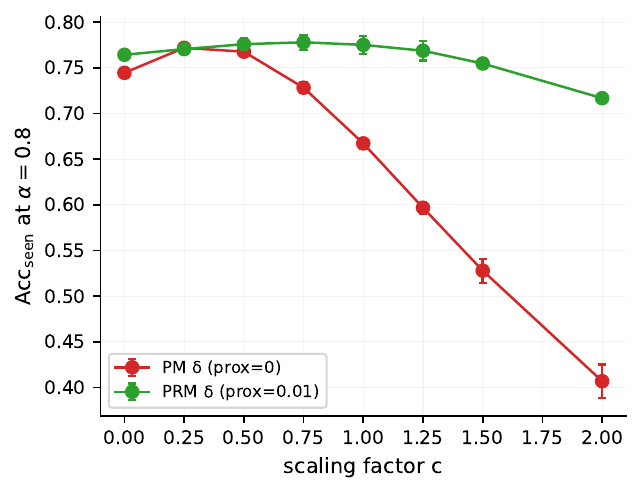}
\caption{$\overline{\mathrm{Acc}}_{\le t}@\alpha{=}0.8$ vs.\ $c$.}
\end{subfigure}
\caption{Continuous task-vector rescaling intervention on CLIP IN-R
10T (task~5).  We sweep
$c\in\{0,\tfrac14,\tfrac12,\tfrac34,1,\tfrac54,\tfrac32,2\}$ for both
P\&M and \prm{} vectors.  The sweep verifies the expected scaling
relations $\|\delta(c)\|_F=c\|\delta(1)\|_F$ and
$q_t(c)=c^2q_t(1)$, while coefficient sensitivity and fixed-$\alpha$
accuracy vary continuously with $c$.}
\label{fig:scaling-intervention}
\end{figure}

We use the name ``continuous task-vector rescaling intervention'' to
make clear that the intervention rescales the task vector, not
the model.  Figure~\ref{fig:scaling-intervention} reports six panels:
the Frobenius norm and Fisher-weighted interference confirm the
expected $\|\delta(c)\|_F=c\|\delta(1)\|_F$ and $q_t(c)=c^2q_t(1)$
scaling, while $R_t$, $W_{t,\varepsilon}$, $G_{t,\mathrm{fix}}$, and
$\overline{\mathrm{Acc}}_{\le t}{@}\alpha{=}0.8$ all vary continuously
with $c$.  Together with the norm-matched swap, this analysis supports
the interpretation that task-vector radius is a major axis behind
the observed coefficient sensitivity.

\section{Additional ablation notes}
\label{app:prox-ablation}
\label{app:prox-sweep}  

The main text already reports the full proximal-strength sweep
(Figure~3) and the component ablation (Table~\ref{tab:ablation}).  We therefore do not duplicate those
figures or tables here.  This appendix only records the interpretation
and the diagnostic caveat for NoWrite.

\subsection{Component-ablation interpretation}
\label{app:ablation}

Table~\ref{tab:ablation} separates perturbation-aware task-vector training,
proximal task-vector shaping, the write-in coefficient rule, and the final write-in.
The key comparisons are: P\&M-fixed vs. \prm{}-fixed isolates the effect
of adding proximity at the same fixed $\alpha$; P\&M vs. \prm{}-Fisher
keeps the Fisher write-in coefficient rule; and LoRA-Prox tests whether the
proximal term already changes mergeability without perturbation.  The
NoWrite rows are attribution probes rather than deployable methods.

\subsection{NoWrite caveat}
\label{app:nowrite-caveat}

NoWrite is purely diagnostic.  Setting $\alpha{=}0$ at the write-in step
means the per-task LoRA factors are still trained under the task-vector-training
objective, but no task vector is permanently accumulated into the
running model.  This isolates task-vector-training capacity before write-in, but it
is not a deployable continual-learning method: it cannot retain new
post-task capacity through the shared running model.  The NoWrite rows
therefore support attribution, not a baseline comparison.

\FloatBarrier

\section{When does \prm{} help most?}
\label{app:applicability}

The mechanism account predicts that proximal shaping should be most
useful when the unregularized task vector induces a sharp or brittle
write-in path.  This appendix should be read as a diagnostic analysis,
not as a predictive law: across the measured settings, larger baseline
coefficient sensitivity is associated with larger \prm{} gains.

\subsection{Baseline brittleness metrics}
\label{app:applicability:metrics}

For a coefficient grid $\mathcal{A}=\{0,0.05,\dots,1.0\}$, we measure
properties of the baseline P\&M path: the write-in range
$\overline{R}_t$ and the fixed-rule gap
$\overline{G}_{t,\mathrm{fix}}$ (Appendix~\ref{app:details}).  Larger
$\overline{R}_t$ means the path is more coefficient-sensitive; larger
$\overline{G}_{t,\mathrm{fix}}$ means a larger cost of deploying the
fixed $\alpha_r{=}0.8$.  For settings without an $\alpha$ sweep, we
report the per-task interference $\overline{q}_t$.

\subsection{Cross-setting relation between mergeability and gain}
\label{app:applicability:scatter}

\begin{table}[H]
\centering
\caption{Cross-setting summary.  $\overline{R}_t$ and
$\overline{q}_t$ are baseline P\&M values; $\Delta\textsc{AAA}$ is
\prm{}-fixed $-$ P\&M; forgetting reduction is P\&M Forg.\ $-$
\prm{}-Fisher Forg.\ (positive means improvement).  ``--'' in
$\overline{R}_t$ marks settings without an $\alpha$ sweep.}
\label{tab:cross-setting}
\resizebox{0.95\textwidth}{!}{%
\setlength{\tabcolsep}{4pt}
\renewcommand{\arraystretch}{1.05}
\begin{tabular}{l l c r r r}
\toprule
Backbone & Setting & $\overline{R}_t^{\textsc{P\&M}}$ [pp] & $\overline{q}_t^{\textsc{P\&M}}$ [$10^{-3}$] & $\Delta\textsc{AAA}$ & Forg.\ reduction \\
\midrule
\multicolumn{6}{l}{\emph{Settings with $\alpha$-sweep instrumentation (R$_t$ available)}} \\
AugReg & IR-10T & 6.78 & 43.74 & +0.84 & -0.78 \\
AugReg & DomainNet 5T & 7.92 & 77.20 & +1.81 & +3.07 \\
AugReg & IR-5T & 7.05 & 120.97 & -0.52 & +0.32 \\
AugReg & IR-20T & 6.88 & 15.12 & +0.54 & -1.28 \\
AugReg & CIFAR-100 10T & 8.46 & 39.92 & +0.81 & -0.16 \\
AugReg & CIFAR-100 20T & 8.47 & 11.54 & -0.32 & -1.22 \\
AugReg & CUB-200 10T & 7.47 & 0.09 & +0.27 & -0.03 \\
CLIP & IR-10T & 20.00 & 147.73 & +6.61 & +1.79 \\
\midrule
\multicolumn{6}{l}{\emph{Settings with per-task instrumentation only (q$_t$; no $\alpha$-sweep)}} \\
CLIP & IR-5T & -- & 412.65 & +6.41 & +5.85 \\
CLIP & IR-20T & -- & 56.89 & +3.06 & -1.16 \\
CLIP & CIFAR-100 10T & -- & 149.44 & +6.05 & +0.77 \\
CLIP & CIFAR-100 20T & -- & 45.37 & +8.66 & -0.67 \\
CLIP & DomainNet 5T & -- & 233.39 & +4.96 & +4.58 \\
CLIP & CUB-200 10T & -- & 4.86 & +3.67 & +0.30 \\
\bottomrule
\end{tabular}
}
\end{table}

\begin{figure}[H]
\centering
\includegraphics[width=0.86\textwidth]{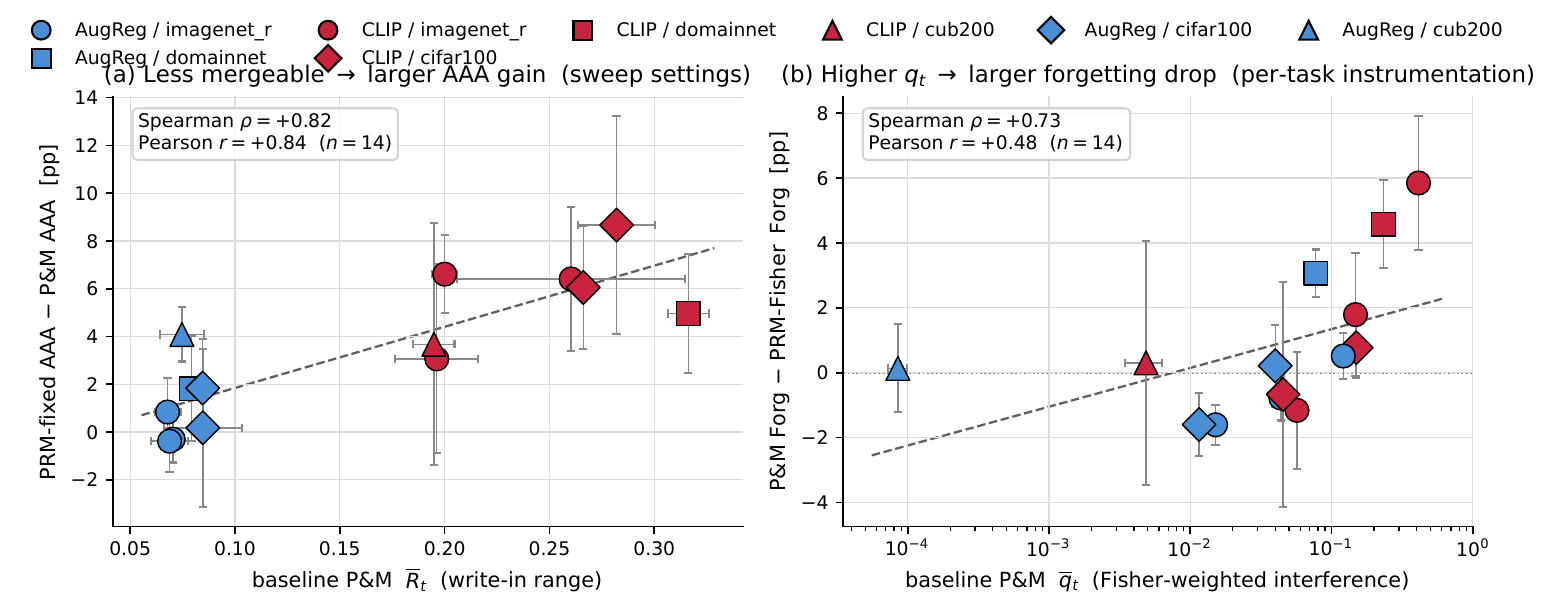}
\caption{Cross-setting gain pattern.  \textbf{(a)} Baseline P\&M
$\overline{R}_t$ vs. $\Delta\textsc{AAA}$ (\prm{}-fixed $-$ P\&M).
\textbf{(b)} Baseline P\&M $\overline{q}_t$ vs. forgetting reduction
(P\&M $-$ \prm{}-Fisher).  Markers encode dataset and backbone.}
\label{fig:appG-gain-scatter}
\end{figure}

\paragraph{Pattern.}
Figure~\ref{fig:appG-gain-scatter} plots the cross-setting relation
between baseline brittleness and \prm{} gain, and
Table~\ref{tab:cross-setting} reports baseline $\overline{R}_t$,
$\overline{q}_t$, $\Delta\textsc{AAA}$ and forgetting reduction for every
measured setting.  Across these settings, high baseline interference and
sharper write-in paths tend to coincide with larger gains, while
near-mergeable settings leave little headroom.  Because the number of
settings with full $\alpha$-sweep instrumentation is small, we treat
these relations as supporting diagnostics rather than as standalone
statistical claims.

\subsection{Practical guidance}
\label{app:applicability:guidance}

A practitioner can use a one-task $\alpha$ sweep on the unshaped
baseline to estimate $\overline{R}_t$ and $\overline{q}_t$ before
committing to a regularizer.  The sweep is rehearsal-free and requires
no training of \prm{}.  In our measurements, sharper baseline write-in
paths and larger fixed-coefficient gaps are the strongest indicators
that proximal shaping will help.

\section{Broader impacts}
PRM is a general method for rehearsal-free continual adaptation. Potential positive impacts include reducing the need to store previous-task data, improving continual adaptation in storage- or privacy-constrained settings, and making sequential LoRA merging less sensitive to per-task coefficient tuning. Potential negative impacts may arise if continual adaptation is deployed in safety-sensitive or socially consequential applications without adequate validation: model updates could preserve or amplify biases in pretrained backbones, and failures or forgetting could affect downstream users. This work does not release high-risk generative models or scraped datasets, and deployment in sensitive domains should require application-specific safety, fairness, and robustness evaluation.

\FloatBarrier

\end{document}